\documentclass[twoside,leqno,twocolumn]{article}
\usepackage[letterpaper]{geometry}

\usepackage{ltexpprt}

\usepackage{amsmath,amssymb,amsfonts}
\usepackage{algorithmic}
\usepackage{graphicx}
\usepackage{textcomp}
\usepackage{xcolor}
\usepackage{bm}
\usepackage{placeins}
\usepackage{multirow}
\usepackage{multicol}
\newcommand\upt{\mathord{\mathrm{t}}}
\newcommand\upr{\mathord{\mathrm{r}}}
\newcommand\ups{\mathord{\mathrm{s}}}
\newcommand\upif{\mathord{\mathrm{if}}}
\newcommand\upotherwise{\mathord{\mathrm{otherwise}}}

\DeclareMathOperator*{\argmin}{arg\,min}

\usepackage{mathtools, nccmath}
\usepackage{lipsum}
\usepackage{booktabs}  
\usepackage{algorithm}
\usepackage{subfigure}
\usepackage{float}
\usepackage{varwidth}

\begin{document}

\title{Unsupervised Selective Manifold Regularized Matrix Factorization}
\author{Priya Mani\thanks{George Mason University, USA. Email: pmani@gmu.edu}
\and Carlotta Domeniconi\thanks{George Mason University, USA. Email: carlotta@cs.gmu.edu}
\and Igor Griva\thanks{George Mason University, USA. Email: igriva@gmu.edu}}

\date{}

\maketitle


\fancyfoot[R]{\scriptsize{Copyright \textcopyright\ 20XX by SIAM\\
Unauthorized reproduction of this article is prohibited}}





\begin{abstract} \small\baselineskip=9pt Manifold regularization methods for matrix factorization rely on the \textit{cluster} assumption, whereby the neighborhood structure of data in the input space is preserved in the factorization space. 
We argue that using the $k$-neighborhoods of all data points as regularization constraints can negatively affect the quality of the factorization,
and propose an unsupervised and selective regularized matrix factorization algorithm to tackle this problem. Our approach jointly learns  a sparse set of representatives and their neighbor affinities, and the data factorization.
We further propose a fast approximation of our approach by relaxing the selectivity constraints on the data. Our proposed algorithms are competitive against baselines and state-of-the-art manifold regularization and clustering algorithms.
\end{abstract}

\section{Introduction.}
\label{intro}
Non-negative matrix factorization (NMF) \cite{NMF} is an  algebraic method with applications to several domains \cite{DOCCLUST2,COLLAB1,MICROARRAY,SIGNALPROCESS}. NMF decomposes a data matrix $\mathbf{X}$ into a product of two factor matrices as follows:
$\argmin_{{\bf{F}} \geq 0,{\bf{G}} \geq 0} ||{\bf{X}}-{\bf{FG}}^T||_{F}^{2}$, 
where $||.||_{F}$ is the Frobenius norm. The factor matrix ${\mathbf{F}}$ gives a set of basis vectors, and ${\mathbf{G}}$ contains the coefficients of the basis vectors. By imposing a non-negativity constraint on the data, NMF  enables an interpretable, \textit{part-based} representation of the data, through an additive combination of the basis vectors. When NMF is interpreted as a clustering problem, ${\bf{F}}$ represents the latent features for each cluster, and ${\bf{G}}$ is the cluster indicator matrix of the data. 

However, standard NMF is performed in the Euclidean space, and does not incorporate the intrinsic geometric structure of the data into the factorization. This deteriorates the performance of NMF when data lie in a manifold embedded in a high-dimensional space. Manifold regularization methods are applied to NMF to address this issue. They rely on the \textit{cluster} assumption \cite{LOCALITY}, which states that nearby points in input space are also close in the embedding space. The data manifold is usually estimated by constructing the $k$-nearest neighbor ($k$-NN) graph of the data, where nodes correspond to points and edge weights represent pairwise affinities.  Manifold regularization was applied to NMF \cite{GMNMF}, where the Laplacian of the $k$-NN graph affinity matrix is incorporated as a regularization term.
The regularizer ensures that the neighborhood structure of the input space is preserved in the factorization space. 

The $k$-neighborhoods used to approximate a data manifold may contain points that belong to different clusters (or classes). For example, the cluster assumption is often violated in neighborhoods close to the boundary between clusters \cite{HUBNESS}. Furthermore, in high-dimensional data, the cluster assumption is  weakened by the emergence of the hubness phenomenon \cite{HUBNESS}. Specifically, bad hubs may emerge, where a bad hub is a frequent nearest neighbor of  points which belong to a different class than the hub's. In other words, bad hubs are frequent (bad) neighbors with mismatched labels. Bad hubs negatively affects the performance of information retrieval and classification (clustering) (e.g., \cite{HUBNESS-VECTORSPACE,HUBS-MUSIC}). 
In clustering, the lack of labels makes the identification of bad hubs a difficult challenge. If the latter remain undetected, the performance of matrix factorization for clustering can deteriorate. 

We argue that performing regularization using all the $k$-neighborhoods  can negatively affect the quality of data factorization. 
As such, in this work, to reduce the number of bad neighbors and their negative effect on clustering, we propose to selectively regularize matrix factorization by learning representative points whose $k$-neighborhoods have a smaller dispersion, measured in terms of pairwise dissimilarities. 

In Table \ref{tab:NOISYNEIGHBORS}, we analyze the degree of violation of the cluster assumption in the $k$-NN graph of simulated and real data (described in Table~\ref{tab:DATASETS}). The two moons data sets (2-dim and 10-dim embedding) consists of 250 points for each of the two clusters. For each data set, $k=\sqrt{n}$, where $n$ is the number of points. 
The first column gives the percentage of label mismatch between each point and its $k$-nearest neighbors (\% Bad NN) in input space.  The second column gives the  percentage of $k$-neighborhoods which have at least 50\% of label mismatch (\% Bad NBH); we call these {\it{bad neighborhoods}}. The third and fourth columns give the average pairwise similarity of good (i.e. not bad) and bad neighborhoods, respectively, measured between a data point and its $k$-nearest neighbors. An adaptive Gaussian kernel is used to compute the pairwise similarities, with the bandwidth for each data point set equal to the distance to its $k$-th nearest neighbor.
Table \ref{tab:NOISYNEIGHBORS} shows a non-negligible label mismatch among nearest neighbors in most data sets, and the mismatch affects many neighborhoods. 
Furthermore, the average pairwise similarity is often smaller for bad neighborhoods than good ones. The last two columns show the reduction in percentage of label mismatch achieved by our approach, Selective Manifold Regularized Matrix Factorization (SMRMF). For each measure, the best value is bold-faced. We see a considerable reduction in \% Bad NN and \% Bad NBH for the large majority of the data.
\begin{table}[t]
\caption{Cluster assumption violation before and after selective manifold regularization. N/A denotes absence of good or bad neighborhoods. }
\label{tab:NOISYNEIGHBORS}
\begin{center}
\begin{scriptsize}
\begin{tabular}{p{1.2cm}p{0.8cm}p{0.8cm}p{0.8cm}p{0.5cm}|p{0.8cm}p{0.8cm}}
& \multicolumn{4}{c|}{{\bf Before Selection}} & \multicolumn{2}{l}{{ \bf After Selection}}\\
Data set & \% Bad & \% Bad  & Sim.  & Sim.  & \% Bad & \% Bad \\
&   NN & NBH & (good  & (bad & NN & NBH \\
& & & NBH) & NBH) & & \\
\midrule
2DMoons  & 0.01 & 0.00 & 0.78 & N/A & {\bf 0.00} & 0.00\\
10DMoons & 0.21 & 0.09 & 0.67 & 0.68 & {\bf 0.05} & {\bf 0.03}\\
Wave & {\bf 0.27} & {\bf 0.14} & 0.01 & 0.01 &  0.51 & 0.65\\
Ionosphere & 0.59 & 0.60 & 0.20 & 0.18 & {\bf 0.17} & {\bf 0.14}\\
Sonar &  0.37 & 0.20 & 0.17 & 0.15 & {\bf 0.27} &  {\bf 0.05}\\
Movement & 0.68 & 0.72 & 0.15 & 0.08 & {\bf 0.27} & {\bf 0.09} \\
Musk1 &  0.37 & 0.38 & 0.17 & 0.13 & {\bf 0.16} & {\bf 0.04} \\
mfeat-fac & 0.53 & 0.52 & 0.04 & 0.02 & {\bf 0.26} & {\bf 0.14}\\
mfeat-pix &  0.23 & 0.11 & 0.26 & 0.29 & {\bf 0.09} & {\bf 0.03} \\
Semeion & {\bf 0.35} & 0.26 & 0.57 & 0.57 & 0.36 & {\bf 0.21}\\
ISOLET & 0.37 & 0.25 & 0.03 & 0.03 & {\bf 0.25} & {\bf 0.06}\\
COIL-5 & {\bf 0.48} & {\bf 0.54} & 0.23 & 0.08 & 0.57 & 0.61\\
20-news & {\bf 0.68} & {\bf 0.82} & 0.76 & 0.75 & 0.72  &  0.83\\
ORL &  0.80 & 1.00 & N/A & 0.05 & {\bf 0.66} & 1.00 \\
OVA\textunderscore Colon & 0.18 & 0.18 & 0.20 & 0.18 & {\bf 0.07} & {\bf  0.00}\\
\end{tabular}
\vspace{-2.5em}
\end{scriptsize}
\end{center}
\end{table}

Several methods have been proposed in sparse optimization to select representatives, called \textit{exemplars}, from the data (e.g., \cite{EXEMPLARS},\cite{CUR}). Here we leverage a sparse optimization technique based on pairwise dissimilarities (DS3) \cite{DS3}, to learn a subset of \textit{exemplars} and their neighborhood affinities. A design consideration of our algorithm is to enable the selective regularization to learn a discriminative data representation in the latent feature space. Our choice of DS3 for sparse optimization enables SMRMF to select neighborhoods with low dispersion, and learn pairwise affinities which reflect the global structure of the data.

\begin{figure}[t]
 \centering
  \subfigure[RMNMF]{
        \centering
        \includegraphics[width=40mm,height=30mm]{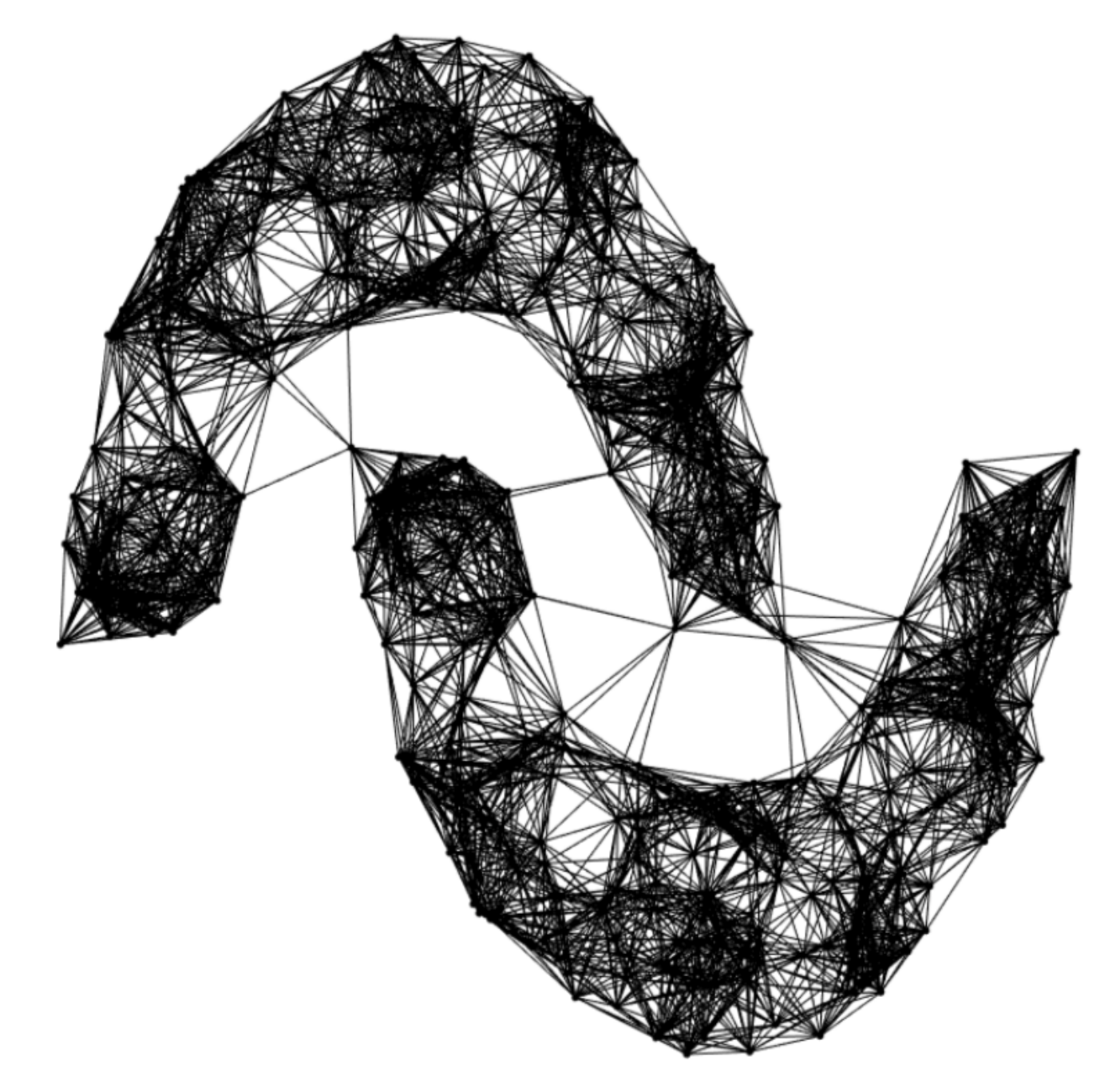}
       }
        \subfigure[SMRMF]{
        \centering
        \includegraphics[width=40mm,height=30mm]{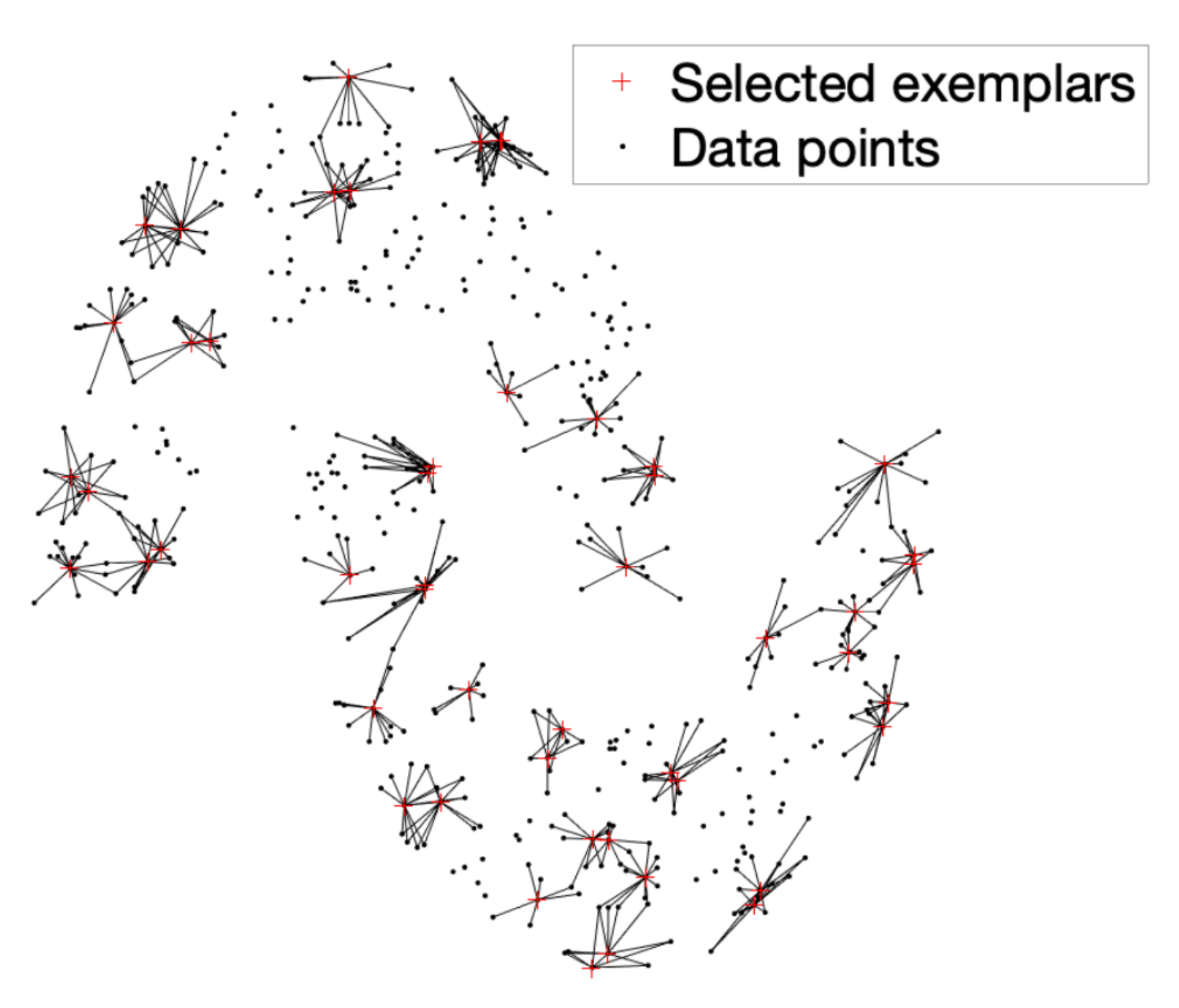}
       }
        \caption{Manifold approximation for 2DMoons.} 
        \vspace{-1.0em}
    \label{fig:2moons}
\end{figure}

To illustrate the result achieved by SMRMF, we show the learned manifold and clustering performance on the 2-moon data in Fig.~\ref{fig:2moons} and Table~\ref{tab:2moons}, respectively. We compare SMRMF against the baseline RMNMF \cite{RMNMF}, which uses all the pairwise affinities in  input space for regularization. SMRMF is run at 10\% exemplar selection for 2DMoons, and at 80\% for 10DMoons.
 SMRMF achieves higher accuracy and NMI than RMNMF on both data sets, and has larger  gains for 10DMoons. From Table \ref{tab:NOISYNEIGHBORS}, we can see that SMRMF significantly reduces the value of \% Bad NN in the learned manifold. 

\begin{table}[t]
\caption{Clustering performance on 2-moons.}
\label{tab:2moons}
\begin{center}
\begin{scriptsize}
\begin{tabular}{lll|ll}
\multirow{2}{*}{Evaluation} &
      \multicolumn{2}{c|} {2DMoons } &
      \multicolumn{2}{c}{10DMoons } \\
      & RMNMF & SMRMF &  RMNMF & SMRMF  \\
      \midrule
ACC & 0.8640 & {\bf 0.8800} & 0.6880 & {\bf 0.9240}\\
NMI & 0.4282 & {\bf 0.5391} & 0.1998 & {\bf 0.6521}\\
\end{tabular}
\end{scriptsize}
\end{center}
\vspace{-2.0em}
\end{table}
The main contributions of our paper are as follows:
(1) We propose a novel algorithm (SMRMF) to jointly learn both a sparse set of exemplars and their neighbor affinities, and data factorization, thereby allowing the exemplar selection and the data factorization to be interdependent; 
(2) SMRMF facilitates a data-driven selection of $k$-neighborhoods with low dispersion, and learns pairwise affinities which reflect the global structure of data, thus enabling a discriminative data representation in the latent feature space; 
(3) In contrast with the existing manifold regularization approaches for standard NMF, we perform  exemplar selection and affinity learning in a kernel-mapped space, while factorizing the data in the input space to preserve the interpretability of NMF;
(4) We develop a fast approximation of SMRMF by relaxing the constraints on the exemplar selection objective, and formulate the learning of sparse pairwise affinities as a convex optimization problem;  
(5) Our empirical evaluation shows that the proposed approaches are competitive against state-of-the-art manifold regularization and clustering algorithms.
    



\section{Related Work.}
\label{sec:RELATED-WORK}
In this section, we briefly discuss related work on manifold regularization and data subset selection.

\textbf{Manifold Regularization.}
The authors of \cite{GMNMF} were the first to introduce a graph-based manifold regularization approach for NMF, and proposed a multiplicative updating algorithm to minimize the NMF loss function. In \cite{RMNMF}, a robust formulation of NMF is proposed which uses the $L_{2,1}$ norm for the loss function, to handle noise and outliers. An augmented Lagrangian Multiplier (ALM) method is derived in \cite{RMNMF} to optimize the regularized robust NMF loss function. The authors of \cite{CAN} proposed a Laplacian-based method to simultaneously learn pairwise data similarities and a clustering of the data. In \cite{AMRMF}, a similar approach is used for adaptive learning and clustering within the NMF framework. In \cite{ADAPTIVEPROJECTIVE-NMF}, the data is projected onto a lower-dimensional subspace; affinity learning and graph clustering are then performed in the projected subspace.  The aforementioned methods use the neighborhoods of all the data, and do not incorporate any global information or selection process in learning the neighborhoods.  

Semi-supervised extensions to manifold regularization \cite{SEMI-NMF, CNMF, GSNMF} utilize label information to generate pairwise constraints to improve the discriminability of classes in a latent space. In \cite{RSNMF}, labels are used to explicitly learn a block-diagonal structure on the latent representation. In \cite{PCOG}, the algorithm takes a set of must-link and cannot-link constraints as input, and learns an optimized graph and clustering which supports the given constraints with guarantees on the cannot-link constraint pairs to be in different clusters. Other variants of manifold regularized NMF include regularization on hyper-graphs, multi-view learning, and deep learning based methods \cite{HYPERGRAPH-NMF,MULTI-MANIFOLD,DEEP-SEMINMF}. 

SMRMF is inspired by the robust loss in \cite{RMNMF} and the adaptive learning in \cite{AMRMF}. However, \cite{RMNMF} does not perform adaptive learning of the pairwise affinities of the manifold. 
SMRMF differs from state of the art adaptive manifold regularization (such as \cite{AMRMF,ADAPTIVEPROJECTIVE-NMF}) in several ways. 
We perform exemplar selection and affinity learning through a kernel-mapping, thereby learning a more discriminative latent representation, while also retaining the interpretability of basis vactors in the latent space. 
Current algorithms learn neighborhood similarity vectors independently for each data point and the learned affinities do not incorporate information about the global structure of the data.
Due to the exemplar selection constraints of SMRMF, and to the nature of our manifold approximation, the learned affinity of a data point to its neighbor also depends on the pairwise affinities between the neighbor and the rest of the data. 
The affinities learned by SMRMF are more informative, as their values provide an ordering of the $k$-nearest neighbors of a data point, as well as their degree of affiliation w.r.t. other representative points. 


\textbf{Data Subset Selection.} Subset selection methods have been studied extensively in the literature. Some approaches find representatives that lie in a low dimensional embedded space (e.g., CUR \cite{CUR}, Rank-revealing QR \cite{QR}). Other approaches, like the algorithm introduced in this paper, optimize the pairwise dissimilarities between points to select a data subset \cite{EXEMPLARS,DS3}. Probabilistic models, such as Determinantal Point Processes (DPP, \cite{DPP}) and its variants, select a diverse subset of instances from the data, by modeling negative correlations derived from pairwise similarities. Iterative Projection and Matching (IPM, \cite{IPM}), a state-of-the-art subset selection method, proposes a fast data selection approach by iteratively selecting samples that capture the maximum information of the data structure, and by projecting on the null space of the remaining data. Subset selection is also inherent to certain clustering algorithms such as $K$-medoids \cite{K-MEDOIDS} and Affinity Propagation \cite{AFFINITY}. A state-of-the-art exemplar-based subspace clustering algorithm (ESC-FFS \cite{ESC-FFS}) introduces a farthest-first search algorithm to iteratively select the least well-represented data point as an exemplar within a sparse representation framework.

\section{Selective Manifold Regularization.}
\label{sec:SMRMF}
The manifold on which the data lies is approximated by the $k$-NN graph of the data. However, in general, the \textit{cluster} assumption is not equally preserved  across all $k$-neighborhoods. Our aim is to improve the factorization of the data  by choosing only the $k$-neighborhoods which best preserve the cluster assumption. To this end, we sparsify the $k$-NN graph by selecting a sparse subset of data points with low dispersion among their neighbors. 

Let ${\mathbf{X}} \in \mathbb{R}^{m \times n}$ denote the data matrix of $n$ points ${\mathbf{x} \in \mathbb{R}^m}$. 
A robust manifold regularized matrix factorization of ${\bf{X}}$ for data clustering was proposed in \cite{RMNMF}, and is defined as follows:
\begin{equation}
\label{eq:RMNMF}
\begin{alignedat}{2}
\argmin_{{\mathbf{F}},{\mathbf{G}}} ||{\bf{X}}-{\bf{FG}}^T||_{2,1} + \lambda \upt \upr ({\bf{G}}^T {\bf{LG}})\\ 
\ups.\upt.~ {\bf{G}}^T{\bf{G}}=\bm{I}, {\bf{G}} \geqslant 0 
\end{alignedat}
\end{equation}
where $tr(.)$ denotes the trace of a matrix, ${\bf{F}} \in \mathbb{R}^{m \times c }$ denotes the latent features for each of the $c$ clusters in the data, ${\bf{G}} \in \mathbb{R}^{n \times c}$ is the cluster indicator matrix, and ${\bm I}$ is the identity matrix. ${\bf{L}}$ is the Laplacian matrix defined as ${\bf{L}} = {\bf{Deg}} -{\bf{W}}$, where ${\bf{W}}$ is the symmetric affinity matrix of the $k$-NN graph of the data, and ${\bf{Deg}}$ is the diagonal degree matrix with ${\bf{Deg}}_{ii} = \sum_j {\bf{W}}_{ij}$.
The $L_{2,1}-$norm is less sensitive to outliers and results in a better factorization of ${\bf{X}}$. The second term in Eq. \eqref{eq:RMNMF} represents the manifold regularization, which is derived in terms of the Laplacian of an affinity matrix constructed from the $k$-NN graph of the data.

We extend the matrix factorization framework in \eqref{eq:RMNMF} to  learn a sparse subset of data points and their pairwise affinities, on which manifold regularization is applied. 
We use DS3 \cite{DS3}, a sparse optimization algorithm based on pairwise dissimilarities, to select exemplars from the data. Given a source set 
$\mathbb{X} \in \mathbb{R}^{m \times M}$ and a target set $\mathbb{Y} \in \mathbb{R}^{m \times N}$, both sampled from {\bf{X}},
and pairwise dissimilarities $\left\{d_{ij}\right \}_{i=1,2,\dots,M}^{j=1,2,\dots,N}$ between points of $\mathbb{X}$ and $\mathbb{Y}$, we find \textit{few} elements of $\mathbb{X}$ that well encode the elements in $\mathbb{Y}$. The objective function of the constrained version of DS3 is:
\begin{equation}
\label{eq:DS3-CONSTRAINED}
\begin{alignedat}{2}
\argmin_{z_{ij}} \sum_{j=1}^{N} \sum_{i=1}^{M}d_{ij}z_{ij}~~~~\ups.\upt.
\\\sum_{i=1}^{M} ||\mathbf{z}_i||_{p} \leq \tau; ~\forall{j},~ \sum_{i=1}^{M}z_{ij} = 1;~ \forall{i,j},~z_{ij} \geqslant 0  
\end{alignedat}
\vspace{-0.5em}
\end{equation}
where $z_{ij}$ is the probability that $\mathbf{x}_i$ is a representative of $\mathbf{y}_j$;  $\mathbf{z}_i = (z_{i1}, \dots, z_{iN})$; and $\tau > 0$ is a parameter which controls the exemplar size.  The objective function in Eq. (\ref{eq:DS3-CONSTRAINED}) denotes the total cost of encoding $\mathbb{Y}$ with exemplars from $\mathbb{X}$. The constraint on the $L_p$ norms of the $\mathbf{z}_i$s induces row sparsity to select \textit{few} exemplar points. Typically, the $L_2$ or $L_{\infty}$ norm is used. When $L_{\infty}$ is used, $\tau$ denotes the desired number of exemplars to be selected.

We incorporate the sparse optimization of Eq. \eqref{eq:DS3-CONSTRAINED} into the matrix factorization framework of Eq. \eqref{eq:RMNMF}, and simultaneously minimize the objectives for matrix factorization and exemplar selection. In our case, $\mathbb{X}$ = $\mathbb{Y}$. The unified objective function is given by:
\begin{equation}
\label{eq:SMRMF}
\begin{alignedat}{2}
\argmin_{{\bf{Z}},{\bf{F}},{\bf{G}}} ||{\bf{X}}-{\bf{FG}}^T||_{2,1} + \lambda \upt \upr ({\bf{G}}^T {\bf{L_ZG}}) + 
\beta tr({\bf{D}}^T{\bf{Z}})  \\ 
\ups.\upt.~||{\bf{Z}}||_{1,p} \leq \tau,~ {\bf{1}}^T{\bf{Z}} = {\bf{1}}^T,  {\bf{Z}} \geqslant 0,  {\bf{G}}^T{\bf{G}}={\bm{I}}, {\bf{G}} \geqslant 0 
\end{alignedat}
\end{equation}
where ${\bf D} \in \mathbb{R}^{n \times n}$ contains pairwise dissimilarities between the points of $\mathbf X$; ${\bf{Z}} \in \mathbb{R}_{+}^{n \times n}$ ($\mathbb{R}_{+}$ is the set of non-negative real numbers) learns the affinities between points and their representatives, and identifies the exemplars;
${\bf{L_Z}}$ = ${\bf Deg}_{\bf{Z}}-\frac{{\bf{Z}}+{\bf{Z}}^T}{2}$, where ${\bf Deg}_{\bf{Z}}$ is a diagonal matrix defined as $({\bf Deg}_{\bf{Z}})_{ii} = \sum_{j} \frac{{\bf Z}_{ji}+ {\bf Z}_{ij}}{2}$.

The optimization problem in Eq. \eqref{eq:SMRMF} is not convex in all three variables. Hence, we solve it using the Augmented Lagrangian method (ALM, \cite{RMNMF}). By introducing the auxiliary variables ${\bf{X}}-{\bf{FG}}^T={\bf{E}}$, ${\bf{G}}={\bf{H}}$, and ${\bf{Z}}={\bf{C}}$, we re-write  Eq. \eqref{eq:SMRMF} as follows
\begin{equation}
\label{eq:AUX}
\begin{alignedat}{2}
\argmin_{{\bf{Z}},{\bf{C}},{\bf{F}},{\bf{G}},{\bf{E}},{\bf{H}}} ||{\bf{E}}||_{2,1} + \lambda \upt \upr ({\bf{G}}^T {\bf{L_Z H}}) \\
+ \beta \upt \upr({\bf{D}}^T {\bf{Z}}) + \beta {\it I_{{\bf{C}} \in S}}\\ \ups.\upt~ {\bf{1}}^T {\bf{Z}} = {\bf{1}}^T, {\bf{Z}} \geqslant 0,  
S=\left\{{\bf{C}}:||{\bf{C}}||_{1,p} \leq \tau \right\}, \\
 {\bf{E}}={\bf{X}}-{\bf{FG}}^T, {\bf{H}}={\bf{G}}, {\bf{C}}={\bf{Z}}, {\bf{G}}^T{\bf{G}}={\bm I}, {\bf{H}} \geqslant 0 
 \end{alignedat}
 \end{equation}
 The constraint $ ||{\bf{Z}}||_{1,p} \leq \tau$ is represented by an indicator function on the set $S$ \cite{CONVEXOPT}, where  $I_{{\bf{C}} \in S} = 0$ when ${\bf{C}} \in S$, else $I_{{\bf{C}} \in S} = \infty$.
 
 The augmented Lagrangian for Eq.~\eqref{eq:AUX} is
\begin{equation}
    \label{eq:ALM}
    \begin{alignedat}{2}
\argmin_{{\bf{Z}},{\bf{F}},{\bf{G}},{\bf{E}},{\bf{H}},{\bf{C}},{\bf{\Lambda}}_1,{\bf{\Lambda}}_2,{\bf{\Lambda}}_3,\mu} ||{\bf{E}}||_{2,1} + \lambda \upt \upr ({\bf{G}}^T {\bf{L_Z H}})  \\
    + \beta tr({\bf{D}}^T {\bf{Z}}) + \beta {\it I_{{\bf{C}} \in S}} 
    + \frac{\mu}{2}||{\bf{X}}-{\bf{FG}}^T-{\bf{E}} + \frac{1}{\mu}{\bf{\Lambda}}_1||_F ^2 \\ +\frac{\mu}{2}||{\bf{G}}-{\bf{H}}+\frac{1}{\mu}{\bf{\Lambda}}_2||_F ^2 +\frac{\mu}{2}||{\bf{Z}}-{\bf{C}} + \frac{1}{\mu}{\bf{\Lambda}}_3||_F ^2 \\
    \ups.\upt. ~{\bf{G}}^T{\bf{G}}={\bm{I}}, {\bf{H}} \geq 0, 
    S=\left\{{\bf{C}}:||{\bf{C}}||_{1,p} \leq \tau \right\},\\
    {\bf{1}}^T {\bf{Z}} ={\bf{1}}^T, {\bf{Z}} \geq 0
    \end{alignedat}
    \end{equation}
 where ${\bf{\Lambda}}_1 \in \mathbb{R}^{m \times n}$, ${\bf{\Lambda}}_2 \in \mathbb{R}^{n \times c}$, and ${\bf{\Lambda}}_3 \in \mathbb{R}^{n \times n}$ are matrices of Lagrangian multipliers, and $\mu$ is the penalty coefficient. We solve Eq.~\eqref{eq:ALM} with alternating updates; in each step, one variable is updated and the remaining ones are kept fixed. 
 
 Our update of ${\bf{Z}}$ differs from the standard ALM procedure in two ways: 
 (1) In each iteration, we map the data vectors represented by the latent features onto a kernel space, and learn $\bf Z$ by computing pairwise dissimilarities in the kernel space (see Section~\ref{sec:SMRMFOPT});
 (2) In each iteration, we use a subset of exemplars, as identified by  $\bf{Z}$, to compute an updated neighborhood graph and its affinity matrix. Specifically, the rows of ${\bf Z}$ (${\bf z}_i$s) are ranked in descending order according to their $L_2$ norm, and the $\tau$ representatives corresponding to the highest $L_2$ norms are selected as exemplars.
 
 Let $R$ be the set of selected representatives. 
 The updated affinity matrix $\Hat{{\bf{Z}}}$ is computed as follows 
\begin{equation}
    \label{eq:Zhat}
    \begin{aligned}
    \Hat{{\bf{Z}}} &= {\bf{M}} \odot {\bf{Z}} \\
    {\bf M}_{ij} &= 
    \begin{cases}
          1 & ({\bf x}_i \in R) \wedge ({\bf x}_j \in k\text{NN}({\bf x}_i))\\
          0 & \text{\it otherwise}
    \end{cases}
    \end{aligned}
    \end{equation}
    where $\odot$ is the Hadamard product. The Laplacian  ${\bf L}_{\bf Z}$ in Eq.~\eqref{eq:ALM} is approximated as  ${\bf{L_{\Hat{Z}}}}$ = ${\bf Deg}_{\Hat{{\bf{Z}}}}-\frac{{\Hat{{\bf{Z}}}}+{\Hat{{\bf{Z}}}}^T}{2}$. Thus, the exemplars identified at each iteration influence the manifold regularization through the Laplacian $L_{\Hat{{\bf{Z}}}}$. In turn, the pairwise dissimilarities between data vectors in the latent space influence the update of ${\bf{Z}}$, where the latent space is defined by the $c$ basis vectors of ${\bf F}$. 

\begin{algorithm}[t]  
\caption{SMRMF}
\label{alg:SMRMF-UNIFIED}
\begin{footnotesize}
\begin{algorithmic}[1]
   \STATE {\bfseries input:} Data ${\bf{X}} \in \mathbb{R}^{m \times n}$, rank $c$, exemplar size $\tau$, ALM parameters $\mu$, $\rho$
   \STATE {\bfseries output:} factor matrices ${\bf{F}}$ and ${\bf{G}}$, encoding matrix ${\bf{Z}}$ \\
   
   \STATE ${\bf{F}}, {\bf{G}}:= {\text{NNSVD}}({\bf{X}})$  
   \STATE ${\bf{D}}$: = computeDissimilarity(${\bf{X}}$)
   \STATE ${\bf{Z}}:$ = A2DM2(${\bf{D}}$)
   \REPEAT
   \STATE $\Hat{{\bf{Z}}}:$ = computeAffinity(${\bf{Z}}$)
   \STATE ${\bf{L_{\Hat{Z}}}}:$ = ${\bf Deg}_{\Hat{\bf{Z}}} - \frac{\Hat{{\bf{Z}}}+\Hat{{\bf{Z}}}^T}{2}$ 
   \STATE  \begin{varwidth}{\linewidth}
   Update ${\bf{E}}$: \par
   \hskip \algorithmicindent ${\bf{e}}_i = 
   \begin{cases}
      (1-\frac{1}{\mu||{\bf{b}}_i||}){\bf{b}}_i, &  \upif~  ||{\bf{b}}_i|| \geq \frac{1}{\mu} \\
      0,        & \upotherwise
    \end{cases}$ \par
    \hskip \algorithmicindent where ${\bf b}_i$ is the $i^{th}$ column of ${\bf{B}}= {\bf{X}}-{\bf{FG}}^T+\frac{1}{\mu}{\bf{\Lambda}}_1$
    \end{varwidth}
  \STATE Update ${\bf{F}}=({\bf{X}}-{\bf{E}}+\frac{1}{\mu}{\bf{\Lambda}}_1){\bf{G}}({\bf G}^T{\bf G})^{-1}$
  \STATE \begin{varwidth}{\linewidth} 
  Update ${\bf{H}}$: \par
  \hskip \algorithmicindent ${\bf{H}}_{ij}= max({\bf{J}}_{ij},0), i=1,2,\dots,n, ~j=1,2,\dots,c.$ \par
  \hskip \algorithmicindent where ${\bf{J}}= ({\bf{G}}+\frac{1}{\mu}{\bf{\Lambda}}_2 -\frac{\lambda}{\mu}{\bf{L_Z}}{\bf G})$
  \end{varwidth}
  \STATE \begin{varwidth}{\linewidth}
  Update ${\bf{G}}={\bf{UV}}^T$: \par
  \hskip \algorithmicindent where ${\bf U}$ and ${\bf V}$ are left and right singular values of \par
  \hskip \algorithmicindent ${\bf{H}} - \frac{1}{\mu}{\bf{\Lambda}}_2 - \frac{\lambda}{\mu}{\bf{L_Z H}} + ({\bf{X}}-{\bf{E}} + \frac{1}{\mu}{\bf{\Lambda}}_1)^T {\bf{F}}$
  \end{varwidth}
   \STATE Compute $\Hat{{\bf{D}}} = {\bf{D}}_K + \frac{\lambda}{\beta}{\bf{D}}_K^{hg}$
   \STATE Update ${\bf{Z}}$ using $\Hat{{\bf{D}}}$ by equation \eqref{eq:dhat}
   \STATE Update ${\bf{C}}$ by equation \eqref{eq:C}
   \STATE \begin{varwidth}{\linewidth} 
   Update  ${\bf{\Lambda}}_1$, ${\bf{\Lambda}}_2$, ${\bf{\Lambda}}_3$, $\mu$ \par
   \hskip \algorithmicindent ${\bf{\Lambda}}_1={\bf{\Lambda}}_1+ \mu({\bf{X}}-{\bf{FG}}^T -{\bf{E}})$ \par
   \hskip \algorithmicindent ${\bf{\Lambda}}_2={\bf{\Lambda}}_2+ \mu({\bf{G}}-{\bf{H}})$ \par
   \hskip \algorithmicindent ${\bf{\Lambda}}_3={\bf{\Lambda}}_3+ \mu({\bf{Z}}-{\bf{C}})$ \par
   \hskip \algorithmicindent $\mu=\rho\mu$ 
   \end{varwidth}
   \UNTIL{convergence}
   \end{algorithmic}
   \end{footnotesize}
   \end{algorithm}

   \section{SMRMF Optimization.}
\label{sec:SMRMFOPT}
Eq. \eqref{eq:ALM} involves several variables and can be solved by alternating minimization, where one variable at a time is updated while the others are kept fixed. The variables are updated using ALM, which is a widely used approach \cite{RMNMF,AMRMF,ADAPTIVEPROJECTIVE-NMF}. The updates for variables ${\bf E}$, ${\bf F}$, ${\bf G}$, and ${\bf H}$ in \eqref{eq:ALM} can be derived using the standard ALM procedure (see supplementary material for the derivations). Algorithm~\ref{alg:SMRMF-UNIFIED} summarizes SMRMF and includes the update equations of all variables.  Below we describe the steps to update ${\bf Z}$ and ${\bf C}$.
\noindent \textbf{Update ${\bf{Z}}$:} When optimizing with respect to ${\bf{Z}}$, Eq. \eqref{eq:ALM} becomes\vspace{-0.5em}
\begin{equation}
\label{eq:Z}
\vspace{-0.5em}\begin{alignedat}{2}
 \argmin_{\bf{Z}}  \lambda tr({\bf{G}}^T{\bf{L_Z}}{\bf{H}}) + \beta tr({\bf{D}}^T{\bf{Z}}) + \\\frac{\mu}{2} || {\bf{Z}}-{\bf{C}}+\frac{{\bf \Lambda}_3}{\mu}||_F^2 ~~\ups. \upt.~ {\bf{1}}^T{\bf{Z}}={\bf{1}}^T, {\bf{Z}}\geq 0 \\
 \end{alignedat}
 \end{equation}
 In order to solve for {\bf Z}, Eq.~\eqref{eq:Z} is re-written in terms of inner products of dissimilarity matrices and {\bf Z}
 \begin{equation}
\label{eq:ZtoD}
\begin{alignedat}{2}
 \argmin_{\bf{Z}} \lambda \sum_{i}\sum_{j}||{\bf g}_i -{\bf h}_j||_2^2{\bf Z} + \beta tr({\bf{D}}^T{\bf{Z}}) + \\\frac{\mu}{2} || {\bf{Z}}-{\bf{C}}+\frac{{\bf \Lambda}_3}{\mu}||_F^2 \\
 = \argmin_{\bf{Z}} \beta \langle{\bf D},{\bf Z}\rangle + \lambda \langle{\bf D}^{hg},  {\bf Z}\rangle +  
 \frac{\mu}{2} || {\bf{Z}}-{\bf{C}}+\frac{{\bf \Lambda}_3}{\mu}||_F^2 \\~\ups. \upt. {\bf{1}}^T{\bf{Z}}={\bf{1}}^T, {\bf{Z}}\geq 0 \\
 \end{alignedat}
 \end{equation}
  where ${\bf g}_i, {\bf h}_j \in \mathbb{R}_{+}^{1\times c}$  are row vectors of $\bf{G}$ and $\bf{H}$ respectively; ${\bf D}_{ij}$ is the dissimilarity between  points ${\bf x}_i, {\bf x}_j$ 
  in the original space; ${\bf D}_{ij}^{hg}$ is the  dissimilarity between ${\bf g}_i$ and ${\bf h}_j$, which are the encodings of ${\bf x}_i$ and ${\bf x}_j$ in the latent space spanned by the $c$ basis vectors~of~${\bf F}$.
 
    We solve the minimization in Eq.~\eqref{eq:ZtoD} in a kernel space. We map the data from the original and latent spaces onto new spaces via Gaussian kernels. The corresponding kernel matrices  ${\bf K}$ and ${\bf K}^{hg}$ are \vspace{-0.5em}
    \begin{equation}
    \label{eq:kernel}\vspace{-0.5em}
    {\bf K}_{ij} = e^{-\frac{1}{2\sigma_i^2}||{\bf{x}}_i - {\bf{x}}_j||^2}~~~
    {\bf K}_{ij}^{hg} = e^{-\frac{1}{2\gamma_i^2}||{\bf{g}}_i - {\bf{h}}_j||^2}
    \end{equation}
    where $\sigma_i$ and $\gamma_i$ are adaptive bandwidths which are set,  for each  point, to the distance to the $k^{th}$ nearest neighbor.
We compute new pairwise dissimilarity matrices ${\bf D}_K$ and ${\bf D}_K^{hg}$ in the respective kernel spaces as follows 
\begin{equation}
    \label{eq:dkernel}
    ({\bf D}_{K})_{ij} = -\frac{({\bf K}_{ij}+ {\bf K}_{ji})}{2}~~~ ({\bf{D}}_{K}^{hg})_{ij} = -\frac{({\bf K}_{ij}^{hg} + {\bf K}_{ji}^{hg})}{2}  
    \end{equation}
  We replace the dissimilarity matrices ${\bf D}$ and ${\bf D}^{hg}$ in Eq.~\eqref{eq:ZtoD} with ${\bf D}_K$ and ${\bf D}_K^{hg}$, and approximate the minimization problem  with the following 
  \begin{equation}
    \label{eq:dnew}
    \begin{alignedat}{2}
 \argmin_{{\bf Z}}\beta\langle{\bf D}_K, {\bf Z}\rangle + \lambda \langle {\bf D}_K^{hg}, {\bf Z}\rangle +  
    \frac{\mu}{2} || {\bf Z}-{\bf C}+\frac{{\bf \Lambda}_3}{\mu}||_F^2
    \\\ups. \upt.~ {\bf 1}^T{\bf Z}={\bf 1}^T, {\bf Z}\geq 0 
    \end{alignedat}
    \end{equation}
The dissimilarity matrices in Eq.~\eqref{eq:dnew} are combined in a unified  matrix $\Hat{{\bf{D}}}_{ij} = ({\bf{D}}_K)_{ij} + \frac{\lambda}{\beta} ({\bf{D}}_K^{hg})_{ij}$. The learning of
    ${\bf{Z}}$ is  based on the dissimilarity values in $\Hat{{\bf{D}}}$, and the minimization problem is formulated as a bounded least squares problem\vspace{-0.5em} \cite{DS3}.
    \begin{equation}
    \label{eq:dhat}
\vspace{-0.5em}    \begin{alignedat}{2}
    \argmin_{\bf{Z}} \beta \langle \Hat{{\bf{D}}}, {\bf{Z}} \rangle +  \frac{\mu}{2} || {\bf{Z}}-{\bf{C}}+\frac{{\bf{\Lambda}}_3}{\mu}||_F^2\\
     =\argmin_{\bf{Z}} \frac{\mu}{2} || {\bf{Z}}-{\bf{C}}+\frac{{\bf{\Lambda}}_3}{\mu} + \frac{\beta}{\mu} \Hat{{\bf{D}}}||_F^2 \\
     \ups. \upt. ~{\bf{1}}^T{\bf{Z}}={\bf{1}}^T, {\bf{Z}}\geqslant 0
    \end{alignedat}
    \end{equation}
SMRMF performs exemplar selection and affinity learning in a kernel space, while the factor matrices {\bf F} and {\bf G} are updated in input space. 
The computation of exemplar selection and affinity learning in a kernel space has multiple advantages: (1) kernel mapping improves  class separability and yields informative dissimilarities to learn exemplars and to regularize the matrix factorization; (2) the updating of factor matrices in input space avoids the kernel pre-image problem \cite{PRE-IMAGE}; (3) each update of the matrix ${\bf Z}$ is performed on a dissimilarity matrix transformed by an adaptive Gaussian kernel with the bandwidth of each point set equal to the distance to its $k^{th}$ nearest neighbor. This operation allows pairwise affinities of farther points to effectively go to zero, thus reducing the number of variables to be updated.

Furthermore, the nature of the learned neighborhoods in SMRMF is different from the existing adaptive manifold regularization methods (e.g. AMRMF \cite{AMRMF}). In fact, the latter learns the pairwise affinities and $k$-neighbors for each data point independently. In contrast, we impose a row-sum constraint on ${\bf Z}$ to learn a distribution over affinities for each column of ${\bf Z}$, while the $k$-NN graph for approximating the manifold are learned from each row of ${\bf Z}$, as shown in \eqref{eq:Zhat}. Therefore, the learned affinity between a data point and its neighbor depends on the distribution of pairwise affinities between the neighbor and the rest of the data. As such, the resulting manifold leads to a discriminative latent representation, as the pairwise affinities are informative of the global structure of the data.

\noindent \textbf{Update ${\bf{C}}$:} When optimizing with respect to ${\bf{C}}$, Eq. \eqref{eq:ALM} becomes\vspace{-0.5em}
\begin{equation}
    \label{eq:C}
    \begin{alignedat}{2}\vspace{-0.5em}
    \argmin_{{\bf{C}}} ~\beta{\it I}_{{\bf{C}} \in S} + \frac{\mu}{2} || {\bf{Z}}-{\bf{C}}+\frac{{\bf{\Lambda}}_3}{\mu}||_F^2\\
    =\argmin_{{\bf{C}} \in S} \frac{1}{2} || {\bf{C}}-{\bf{Z}}-\frac{{\bf{\Lambda}}_3}{\mu}||_F^2\\
    \ups. \upt. ~S=\left\{{\bf{C}}:||{\bf{C}}||_{1,p} \leq \tau \right\}\\
    \end{alignedat}
    \end{equation}
    which can be solved by a projection on $S$. 

The factor matrices ${\bf{F}}$ and ${\bf{G}}$ are initialized by non-negative SVD \cite{NNSVD}. ${\bf{C}}$ is initialized to a single exemplar which has the minimum sum of pairwise distances to other points, where the entries corresponding to its row are set to one. ${\bf{Z}}$ is initialized by optimizing Eq.~\eqref{eq:DS3-CONSTRAINED} using accelerated ADMM (A2DM2) \cite{A2DM2}. A2DM2 requires each of the components in its objective function to be strongly convex. The objective function components in Eq. \eqref{eq:DS3-CONSTRAINED} are not strongly convex, hence we add the squared Frobenius norm of ${\bf{Z}}$ and ${\bf{C}}$ to Eq.~\eqref{eq:DS3-CONSTRAINED} to achieve strong convexity. Hence, we use the following objective function to initialize ${\bf{Z}}$ using A2DM2:
\begin{equation}
\vspace{-0.5em}
\label{eq:DS3-A2DM2}
\begin{alignedat}{2}
\argmin_{{\bf{Z}},{\bf{C}}} tr({\bf{D}}^T{\bf{Z}}) + {\it I_{{\bf{C}} \in S}} + \frac{\delta}{2} ||{\bf{Z}}||_2^2 + \frac{\delta}{2} ||{\bf{C}}||_2^2 \\ \ups.\upt.~ {\bf{Z}}={\bf{C}}, ||{\bf{Z}}||_{1,p} \leq \tau, {\bf{1}}^T {\bf{C}} = {\bf{1}}^T, {\bf{C}} \geqslant 0 \\
S=\left\{{\bf{C}}:||{\bf{C}}||_{1,p} \leq \tau \right\} 
\end{alignedat}
\end{equation}
where $\delta$ is set to a very small value. 
The Augmented Lagrangian for Eq.~\eqref{eq:DS3-A2DM2} is
    \begin{equation}
    \label{eq:Z2CALM}
    \begin{alignedat}{2}
    L({\bf{Z}},{\bf{C}},{\bf{\Lambda}}) = \argmin_{\bf{Z},\bf{C}}  tr({\bf{D}}^T {\bf{Z}}) + {\bf{\it I}}_{{\bf{C}} \in S} + \frac{\delta}{2} || {\bf{Z}} ||_F^{2} \\ + \frac{\delta}{2} || {\bf{C}} ||_F^{2} + \langle {\bf{\Lambda}}, {\bf{Z}} - {\bf{C}} \rangle + \frac{\mu}{2} || {\bf{Z}} - {\bf{C}} ||_F^2 \\ ~\ups.\upt. ~{\bf{1}}^T {\bf{Z}} = {\bf{1}}^T; {\bf{Z}} \geqslant 0; S=\left\{ {\bf{C}} : ||{\bf{C}}||_{1,p} <= \tau \right\}
    \end{alignedat}
    \end{equation}The pseudo code of A2DM2 applied to Eq.~\eqref{eq:Z2CALM} is given in the supplementary material. 
 
 \section{Fast Approximation of SMRMF.}
 Solving \eqref{eq:SMRMF} is computationally expensive due to the $L_{1,p}$ norm constraint on the encoding matrix ${\bf Z}$ for exemplar selection. 
 To speed-up the computation, we develop a fast approximation of SMRMF (f-SMRMF). 
 We relax the constraint $||{\bf Z}||_{1,p}$ in \eqref{eq:SMRMF} to a Frobenius norm $||{\bf Z}||_F^2$, and formulate a unified objective function combining the NMF loss with the relaxed exemplar selection objective. This allows the formulation of a convex optimization problem, which can be easily parallelized to solve each column of ${\bf Z}$ independently. The unified objective function of f-SMRMF is: 
 \begin{equation}
\label{eq:FastSMRMF}
\begin{alignedat}{2}
\argmin_{{\bf{Z}},{\bf{F}},{\bf{G}}} ||{\bf{X}}-{\bf{FG}}^T||_{2,1} + \lambda \upt \upr ({\bf{G}}^T {\bf{L_ZG}}) + \\
\beta (tr({\bf{D}}^T{\bf{Z}}) + \frac{\delta}{2}||{\bf{Z}}||_F^2)  \\ 
\ups.\upt. ~{\bf{1}}^T{\bf{Z}} = {\bf{1}}^T,  {\bf{Z}} \geqslant 0,  {\bf{G}}^T{\bf{G}}={\bm{I}}, {\bf{G}} \geqslant 0 
\end{alignedat}
\end{equation}
We use the procedure in Section~\ref{sec:SMRMFOPT} to solve \eqref{eq:FastSMRMF}. The optimization only differs in the update of ${\bf Z}$, replacing both the updates of ${\bf Z}$ and ${\bf C}$ in Section~\ref{sec:SMRMFOPT}. 
Following steps similar to \eqref{eq:Z}--\eqref{eq:dhat}, the update for ${\bf Z}$ becomes:
\begin{equation}
\label{eq:FastSMRMF-Z}
\begin{alignedat}{2}
\argmin_{{\bf{Z}}}  \langle \Hat{{\bf{D}}}, {\bf{Z}} \rangle +  \frac{\delta}{2} || {\bf{Z}} ||_F^2\\ 
\ups.\upt. ~{\bf{1}}^T{\bf{Z}} = {\bf{1}}^T,  {\bf{Z}} \geqslant 0
\end{alignedat}
\end{equation}
We exploit the fact that each column ${\bf z}_j$ of ${\bf Z}$ in (\ref{eq:FastSMRMF-Z})  can be found independently, by solving the following problem for each column $\Hat{{\bf d}}_j$ of the matrix $\Hat{{\bf D}}$: 
 \begin{equation}
\label{eq:SMRMF-NoSelectBCLS-a}
\vspace{-0.5em}
\begin{alignedat}{2}
\argmin_{{{\bf z}_j}} ~{\Hat{\bf d}_j}^T{{\bf z}_j} + \frac{\delta}{2}||{{\bf z}_j}||_2^2  \\ 
\ups.\upt. ~{\bf{1}}^T{{\bf z}_j} = {{\bf 1}},  {{\bf z}_j} \geqslant 0
\end{alignedat}
\end{equation}
Problem (\ref{eq:SMRMF-NoSelectBCLS-a}) is convex and can be worked out by explicitly solving a Karush-Kuhn-Tucker (KKT) system of equations and inequalities. The time complexity of the resulting algorithm (given in the supplementary material) is $O(n^2)$, where $n$ is the number of instances and the number of rows of ${\bf{Z}}$. Thus, the  cost of computing ${\bf Z}$ in \eqref{eq:FastSMRMF-Z} is $O(n^3)$.
The time complexities to solve \eqref{eq:dhat} and \eqref{eq:C} are $O(n^3)$ and $O(n^2\log(n) +n^3)$, respectively. Thus, f-SMRMF replaces the two update steps for ${\bf Z}$ and ${\bf C}$ with a new single update step for ${\bf Z}$. This gives significant computational gains in the iterative optimization, as also reflected in the empirical running times reported in Section \ref{experiments}.

Due to the Frobenius norm relaxation on ${\bf Z}$, we need to explicitly rank and select exemplars from ${\bf Z}$ (in contrast, the $L_{1,p}$ norm constraint sets the rows of the non-exemplars to zero).  We select the required number $\tau$ of exemplars from a ranking of the rows of ${\bf Z}$, ordered by their descending $L_2$ norms.

 \section{Empirical Evaluation.}
\label{experiments}
We evaluate SMRMF using the data summarized in Table~\ref{tab:DATASETS} (further details on the data are in the supplementary material).  
We compare our approach against several algorithms: baseline matrix factorization and manifold regularization (NMF \cite{NMF}, GNMF \cite{GMNMF}, RMNMF \cite{RMNMF}), baseline subset selection (DS3 \cite{DS3},  DPP \cite{DPP}), state of the art subset selection (IPM \cite{IPM})  state of the art subspace clustering (ESC-FFS, \cite{ESC-FFS}), deep matrix factorization (DSMF, \cite{DEEP-SEMINMF}), and state of the art manifold regularization methods (AMRMF \cite{AMRMF}, APMF \cite{ADAPTIVEPROJECTIVE-NMF}). We  compare two variants of SMRMF: (1) SMRMF-Euc, which learns ${\bf Z}$ in Eq.~\eqref{eq:ZtoD} using pairwise Euclidean distances in the input space, to assess the effect of kernel mapping; (2) SMRMF-NS, which learns ${\bf Z}$ without imposing the exemplar selection constraints. 
\begin{table}[t]
\vspace{-0.7em}
\caption{Summary of the data sets.}
\vspace{-0.5em}
\label{tab:DATASETS}
\vskip 0.15in
\begin{center}
\begin{scriptsize}
\begin{tabular}{llll}
 Data set & \# instances & \# dimensions & \# classes  \\
 \midrule
 Wave & 600 & 21 & 3 \\
 Ionosphere & 351 & 34 & 2 \\
 Sonar & 208 & 60 & 2 \\
 Movement & 360 & 90 & 15 \\
 Musk1 & 476 & 168 & 2 \\
 mfeat-fac & 2000 & 216 & 10 \\
 mfeat-pix &  2000 &  240 & 10 \\
 Semeion & 1593 & 256 & 10 \\
 ISOLET & 1200 & 617 & 5 \\
 COIL-5 & 360 & 1024 & 5 \\
 20-news & 5020 & 1796 & 20 \\
 ORL & 400 & 4096 & 40 \\
 OVA\textunderscore Colon & 1545 & 10935 & 2 \\
\end{tabular}
\end{scriptsize}
\end{center}
\vspace{-2.5em}
\end{table}  
The $k$-neighborhoods of the points selected by DPP and IPM are used to 
selectively regularize RMNMF. The factor matrices for manifold regularization algorithms are initialized using non-negative SVD \cite{NNSVD}. SMRMF and variants are evaluated at 10\% exemplar selection. We tune the regularization parameters of the algorithms 
in the range $\left\{10^{-3}, 10^{-2}, \dots, 1\right\}$ and report the best results. Detailed parameter settings of the algorithms are given in the supplementary material.

\vspace{-0.5em}
\subsection{Evaluation.}
\label{sec:EVALUATION}
 We evaluate the approaches using clustering accuracy (ACC), Normalized Mutual Information (NMI), and running time. 
Tables~\ref{tab:ACC} and \ref{tab:NMI} give the clustering accuracy and NMI of the compared algorithms. NMF, GNMF, DPP, and ESC-FFS are initialized randomly; hence, their results are averaged across 30 iterations. 
The standard deviations of ACC and NMI of DPP are very low (of the order of 1e-15) for all the datasets, hence they are not reported. The initialization and solution of all the other methods are deterministic. DPP on 20-news and OVA\textunderscore Colon could not be run to completion, and APMF could not run to completion on OVA\textunderscore Colon.
\begin{table*}[t]
\caption{Clustering accuracy (ACC) of all methods (standard deviations are given in parentheses).}
\label{tab:ACC}
\begin{center}
\vskip -0.15in
\begin{tiny}
\begin{tabular}{p{1.1cm}lllp{0.5cm}lp{0.5cm}p{0.5cm}p{0.5cm}llllll}
 Data & NMF & DS3 & GNMF &  RMNMF & AMRMF & APMF &  DPP & IPM  & ESC-FFS & DSMF & SMRMF & SMRMF &  SMRMF & f-SMRMF \\
& & & & & & &  & & & &-Euc & -NS  & &  \\
 \midrule
Wave & 0.55 (0.03) & 0.70 & 0.56 (0.04) & 0.71 & 0.78 & 0.63 & 0.71 & 0.74 & 0.53 (0.02) & 0.49 & 0.82 & 0.81 &{\bf 0.85} &  0.84 \\
 Ionosphere & 0.61(0.01) & 0.61 & 0.61 (0.01) & 0.64 & 0.65 & {\bf 0.69} & 0.64 & 0.66 & 0.56 (0.05) & {\bf 0.69}  & 0.66 & 0.65& 0.66 &  0.65 \\
Sonar & 0.54 (0.004) & 0.56 & 0.55 (1e-4) & 0.56 & 0.60 & 0.60 & 0.56 & 0.58 &  0.53 (0.02) & {\bf 0.69} & 0.61 & 0.55 & 0.58 & 0.59  \\
 Movement & 0.32 (0.02) & 0.34 & 0.32 (0.02) &  0.47 & 0.50 & 0.36 & 0.48 & 0.50 & 0.48 (0.01) & 0.35 & {\bf 0.52} & 0.48 & 0.51 &  {\bf 0.52} \\
 Musk1 & 0.53 (1e-15) & 0.54 & 0.53 (1e-15)  & 0.55 & 0.58 & 0.56 & 0.54 & 0.58 & 0.53 (0.01) & 0.54 & 0.57 & 0.57  & {\bf 0.59} &  0.58 \\
 mfeat-fac & 0.51 (0.05) & 0.27 & 0.51 (0.05) & 0.68 & 0.66 & 0.35 & 0.62 & 0.65 & {\bf 0.86}(0.06) & 0.76 & 0.63 & 0.60 & 0.68 &  0.71 \\
 mfeat-pix &  0.43 (0.05) & 0.32 & 0.44 (0.04) & 0.60 & 0.57 & 0.35  & 0.60  & 0.60 & {\bf 0.88} (0.05) & 0.71 & 0.60 & 0.54  & 0.63 &  0.66 \\
Semeion & 0.38 (0.02) & 0.34  & 0.39 (0.02) & 0.50 & 0.48 & 0.23 & 0.53 & 0.49 & {\bf 0.56} (0.03) & 0.48 & 0.47 & 0.42 & 0.50 & 0.53   \\
 ISOLET & 0.56 (0.03) & 0.33 & 0.57 (0.03) & 0.44  & 0.61 & 0.27 & 0.55 & 0.56  & 0.51 (0.004) & 0.51 & 0.64 & 0.49 &  0.61 & {\bf 0.65}  \\
 COIL-5 & 0.77 (0.01) & 0.33 & 0.77 (0.01) & 0.86 & 0.86 & 0.40 & 0.82 & 0.86 &  0.87 (0.10) & 0.58 & 0.86 & 0.84 &  {\bf 0.90} & 0.86  \\
 20-news & 0.26 (0.01) & 0.17 & 0.26 (0.01) & 0.27 & 0.17 & 0.06 & -  & 0.29 & {\bf{0.34}} (0.02) & 0.09 & 0.24 & 0.29 &  0.24 &  0.27 \\
 ORL & 0.40 (0.02) & 0.31 & 0.40 (0.02) &  0.63 & 0.67 & 0.18 &  0.60 &  0.64 & 0.48 (0.02) & 0.17  & 0.64 & 0.60 & 0.67 & {\bf 0.68}  \\
 OVA\textunderscore Colon & 0.55 (1e-4) & 0.79 & 0.81 (1e-15) & 0.87 & 0.88 & - & - & 0.92  & 0.77 (0.0002) & 0.54  & 0.87 & 0.83 & {\bf 0.95} &  {\bf 0.95} \\
 \hline
AVG & 0.49 & 0.43 & 0.52 & 0.60 & 0.61 & 0.39 & 0.60  & 0.62 & 0.61 &  0.51 & 0.62 & 0.59 & 0.64 & {\bf 0.65}\\
\end{tabular}
\end{tiny}
\end{center}
\vspace{-1em}
\end{table*}

\begin{table*}[t]
\vskip -0.15in
\caption{Normalized Mutual Information (NMI) of all methods (standard deviations are given in parentheses).}
\label{tab:NMI}
\vskip -0.1in
\begin{center}
\vskip -0.15in
\begin{tiny}
\begin{tabular}{p{1.1cm}lllp{0.5cm}lp{0.5cm}p{0.5cm}p{0.5cm}llllll}
 Data & NMF & DS3 & GNMF &  RMNMF & AMRMF & APMF &  DPP & IPM  & ESC-FFS & DSMF & SMRMF & SMRMF &  SMRMF & f-SMRMF \\
& & & & & & &  & & & &-Euc & -NS  & &  \\
 \midrule
Wave & 0.38 (0.007) & 0.38 & 0.38 (0.01) & 0.38 & 0.50 & 0.28 & 0.37 & 0.45 & 0.35 (0.01) & 0.21 & 0.50 & 0.49 & {\bf 0.52} & {\bf 0.52}\\
 Ionosphere & 0.02 (0.006) & 0.01 & 0.02 (0.01) & 0.06 & 0.01 & 0.14 & 0.16 & 0.14 & 0.02 (0.02) & 0.07  & 0.17  & 0.16 & 0.09 & {\bf 0.17}\\
 Sonar & 0.004 (0.001) & 0.01 & 0.01 (1e-4) & 0.01 & 0.04 & 0.06 & 0.01 & 0.02 & 0.004 (0.01) & {\bf 0.11}  & 0.03 & 0.005 & 0.02 & 0.02\\
 Movement & 0.37 (0.03) & 0.41 & 0.37 (0.03) & 0.59 & 0.58 & 0.39 & 0.60 & {\bf 0.61} & 0.60 (0.01) & 0.41 & 0.58 & 0.60 & 0.59 & 0.58\\
 Musk1 & 0.01 (1e-15) & {\bf 0.02} & 0.01 (1e-15) & 0.01 & 0.01 & 0.002 & 0.01 & 0.01 & 0.002 (0.001)  & 0.013 & 0.003  & 0.003 & 0.01 & 0.01\\
 mfeat-fac & 0.50 (0.03) & 0.25 & 0.50 (0.04) & 0.58 & 0.61 & 0.31 & 0.56 & 0.57 & {\bf 0.80} (0.02) & 0.72 & 0.58 & 0.55 & 0.60 & 0.60\\
 mfeat-pix & 0.42 (0.04) & 0.30  & 0.42 (0.04) & 0.59 & 0.51 & 0.28 & 0.59 & 0.59 & {\bf 0.82} (0.02) & 0.65 & 0.56 &  0.55 &  0.57 & 0.61\\
Semeion & 0.34 (0.01) & 0.29 & 0.34 (0.01) & 0.41 & 0.42 &  0.11 & 0.45 & 0.43 & {\bf 0.54} (0.02) & 0.42 & 0.40 & 0.29 & 0.42 & 0.44  \\
 ISOLET & 0.45 (0.03) & 0.10  & 0.45 (0.03) & 0.34 & 0.46 & 0.11 & 0.47 & 0.42 & 0.55 (0.01) & {\bf 0.57}   & 0.49 & 0.39 & 0.46 & 0.50\\
 COIL-5 & 0.73 (0.008) & 0.32 & 0.73 (0.01) & 0.84 & 0.84 & 0.25 & 0.77 & 0.84 & 0.85 (0.06) & 0.66  &  0.84 & 0.81 & {\bf 0.88} & 0.84\\
 20-news & 0.25 (0.009) & 0.15 & 0.25 (0.01) & 0.26 & 0.12 & 0.01 & - & 0.27 & {\bf 0.32} (0.01) & 0.04 &  0.24 & 0.29 & 0.23 & 0.27\\
 ORL & 0.61 (0.02) & 0.42 & 0.61 (0.02) & 0.77 & 0.80 & 0.32 & 0.76 & 0.77 & 0.69 (0.01) & 0.30  &  0.78 & 0.76 & {\bf 0.81} & 0.80\\
 OVA\textunderscore Colon & 0.06 (1e-4) & 0.01 & 0.00 (0.00) & 0.30 & 0.27 & - & -  & 0.47 & 0.01 (5e-3) & 0.03   & 0.30 & 0.20 & {\bf 0.60} & {\bf 0.60}\\
 \hline
AVG & 0.32 & 0.20 & 0.31 &  0.39 & 0.40 & 0.19 & 0.43  & 0.43  & 0.43 & 0.32 & 0.42  & 0.39 & 0.45 & {\bf 0.46} \\
\end{tabular}
\end{tiny}
\end{center}
\vspace{-1em}
\end{table*}  

\begin{table*}[t]
\caption{Average running times (in seconds) of all methods.}
\label{tab:RUNTIME}
\begin{center}
\vskip -0.1in
\begin{scriptsize}
\begin{tabular}{llllllllllllll}
 NMF & DS3 & GNMF & RMNMF & AMRMF & APMF & DPP & IPM & ESC-FFS & DSMF & SMRMF & SMRMF & SMRMF & f-SMRMF \\
 & & & & & &  & & & &-Euc & -NS  & &   \\
 \midrule
{\bf 3.02} & 22.99 & 3.96 & 9.78 & 67.23 & 872.64 & 8.64 & 256.50 & 3.71 & 279.23 & 91.91 & 148.45 &  292.47 & 95.77\\
\end{tabular}
\end{scriptsize}
\end{center}
\vspace{-1em}
\end{table*}  


Both SMRMF and f-SMRMF achieve the best average ACC and NMI values, with a considerable margin of improvement on average against all competitors. This demonstrates that our approach is robust under a variety of conditions. In particular, a large improvement margin was achieved by both SMRMF and f-SMRMF on the data with the largest dimensionality (OVA\_Colon). SMRMF and f-SMRMF outperform both variants SMRMF-Euc and SMRMF-NS in average ACC and NMI. This confirms that kernel mapping is effective in learning informative pairwise affinities, and that some $k$-neighborhoods can be detrimental to clustering, and our approach is effective in eliminating them.

Table~\ref{tab:RUNTIME} gives the average running time (in seconds) of the algorithms. The running times for each data set are given in the supplementary material. 
All experiments are run using MATLAB 2019b, on a 2.3 GHz Intel Core i5 processor with 16 GB RAM. NMF is the fastest, but results in poor clustering quality due to the lack of regularization. APMF is the most expensive. f-SMRMF effectively speeds-up SMRMF, while preserving clustering quality, thus making our approach competitive also in terms of complexity. 

\section{Analysis.}
\label{sec:ANALYSIS}

To investigate the effect of our selection procedure, we compare the characteristics of the selected neighborhoods against those discarded.
Table~\ref{tab:EXEMPLARKNN} shows
the average label mismatch of neighbor pairs within the neighborhoods of the selected ({\it{S}}NBH) and non-selected ({\it{NS}}NBH)  points, for SMRMF.
 The third column gives the difference between the two measures. A positive difference indicates that the neighborhoods of the selected  points have lower label mismatch on average than the rest of the  points.
 We observe a positive difference for the majority of the data sets. Furthermore, Tables~\ref{tab:ACC} and \ref{tab:NMI} show that SMRMF achieves a considerable improvement in performance compared to RMNMF in these data sets, thereby demonstrating the effectiveness of our data selection process.
\begin{table}[t]
\vskip -0.15in
\caption{Characterizing the learned neighborhoods.}
\label{tab:EXEMPLARKNN}
\begin{center}
\begin{scriptsize}
\begin{tabular}{llll}
 Data set & AVG Label  &  AVG Label & \\
    & Mismatch  & Mismatch  &  ({\it{NS}}NBH - {\it{S}}NBH)\\
    & ({\it{S}}NBH) & ({\it{NS}}NBH) & \\
 \midrule
 Wave & 0.52 & 0.61 & {\bf 0.09}\\
 Ionosphere & 0.22 & 0.45 & {\bf 0.23} \\
 Sonar &  0.28 & 0.36 & {\bf 0.08}\\
 Movement &  0.22 & 0.31 &  {\bf 0.09}\\
 Musk1 & 0.08 & 0.17 & {\bf 0.09}\\
 mfeat-fac & 0.24 & 0.24 & 0.00 \\
 mfeat-pix & 0.07 & 0.08  & {\bf 0.01} \\
 Semeion & 0.36 & 0.47 & {\bf 0.11}\\
 ISOLET &  0.11 & 0.26 & {\bf 0.15}\\
 COIL-5 & 0.46 & 0.20 & -0.26\\
 20-news & 0.72 & 0.74 & {\bf 0.02}\\
 ORL & 0.70 & 0.73 & {\bf 0.03}\\
 OVA\textunderscore Colon & 0.05 & 0.05 & 0.00  \\
\end{tabular}
\end{scriptsize}
\end{center}
\vspace{-2.5em}
\end{table}  

As highlighted in Table~\ref{tab:NOISYNEIGHBORS}, both \% Bad NN and \% Bad NBH are reduced significantly after exemplar selection, for the majority of the data sets.
 Few data sets (Wave, Semeion, and 20-news) show a rise in \% Bad NN after exemplar selection. This is likely the reason why SMRMF does not outperform RMNMF on Semeion and 20-news. Nevertheless, SMRMF still has a significant improvement over RMNMF on Wave. A plausible explanation is that the affinities learned for the bad neighbors on Wave were not large enough to influence the NMF regularization. Another factor which affects the learned affinities is the separability of data in the latent space learned in the NMF iterations.  We'll further investigate this phenomenon in our future work.
We provide further analysis on the sensitivity of exemplar and neighborhood sizes, and on the convergence of SMRMF, in the supplementary material.
\section{Conclusion.}
We presented a novel algorithm to simultaneously select a subset of $k$-neighborhoods and learn an affinity matrix, within an MF framework. We empirically analyzed the performance of our algorithm and showed that our approach is competitive against a variety of related methods. In the future, we'll investigate a kernelized version of our proposed unified objective function, where the factor matrices are updated in a kernel space.


\end{document}


\title{Unsupervised Selective Manifold Regularized Matrix Factorization \\ Supplementary Material}
\author{Priya Mani\thanks{George Mason University, USA.}
\and Carlotta Domeniconi\thanks{George Mason University, USA.}
\and Igor Griva\thanks{George Mason University, USA.}}

\date{}

\maketitle


\fancyfoot[R]{\scriptsize{Copyright \textcopyright\ 20XX by SIAM\\
Unauthorized reproduction of this article is prohibited}}

\setcounter{algorithm}{1}
\section{SMRMF Optimization.}
We describe below the detailed steps to update variables ${\bf E}$, ${\bf F}$, ${\bf H}$ and ${\bf G}$ in the unified optimization. The Augmented Lagrangian for SMRMF objective (equation (3.5) in the paper) is
\begin{equation}
    \label{eq:ALM}
    \begin{alignedat}{2}
\argmin_{{\bf{Z}},{\bf{F}},{\bf{G}},{\bf{E}},{\bf{H}},{\bf{C}},{\bf{\Lambda}}_1,{\bf{\Lambda}}_2,{\bf{\Lambda}}_3,\mu} ||{\bf{E}}||_{2,1} + \lambda \upt \upr ({\bf{G}}^T {\bf{L_Z H}})  \\
    + \beta tr({\bf{D}}^T {\bf{Z}}) + \beta {\it I_{{\bf{C}} \in S}} 
    + \frac{\mu}{2}||{\bf{X}}-{\bf{FG}}^T-{\bf{E}} + \frac{1}{\mu}{\bf{\Lambda}}_1||_F ^2 \\ +\frac{\mu}{2}||{\bf{G}}-{\bf{H}}+\frac{1}{\mu}{\bf{\Lambda}}_2||_F ^2 +\frac{\mu}{2}||{\bf{Z}}-{\bf{C}} + \frac{1}{\mu}{\bf{\Lambda}}_3||_F ^2 \\
    \ups.\upt. ~{\bf{G}}^T{\bf{G}}={\bm{I}}, {\bf{H}} \geq 0, 
    S=\left\{{\bf{C}}:||{\bf{C}}||_{1,p} \leq \tau \right\},\\
    {\bf{1}}^T {\bf{Z}} ={\bf{1}}^T, {\bf{Z}} \geq 0
    \end{alignedat}
    \end{equation}
 where ${\bf{\Lambda}}_1 \in \mathbb{R}^{m \times n}$, ${\bf{\Lambda}}_2 \in \mathbb{R}^{n \times c}$, and ${\bf{\Lambda}}_3 \in \mathbb{R}^{n \times n}$ are matrices of Lagrangian multipliers, and $\mu$ is the penalty coefficient. \\ 
\noindent \textbf{Update ${\bf{E}}$:} We fix all the variables but ${\bf{E}}$, and remove the terms that do not depend on ${\bf{E}}$. Eq. \eqref{eq:ALM} becomes
\begin{equation}
\label{eq:o3}
\begin{alignedat}{2}
\argmin_{{\bf{E}}} \frac{1}{\mu}||{\bf{E}}||_{2,1} + \frac{1}{2}||{\bf{X}}-{\bf{FG}}^T-{\bf{E}}+ \frac{1}{\mu}{\bf{\Lambda}}_1||_F ^2  
\end{alignedat}
\end{equation}
Let ${\bf{B}}= {\bf{X}}-{\bf{FG}}^T+\frac{1}{\mu}{\bf{\Lambda}}_1$. ${\bf{E}}$ is updated as:
\begin{equation}
\label{eq:s15}
\begin{alignedat}{2}
{\bf{e}}_i = 
\begin{cases}
      (1-\frac{1}{\mu||{\bf{b}}_i||}){\bf{b}}_i, &  \upif~  ||{\bf{b}}_i|| \geq \frac{1}{\mu} \\
      0,        & \upotherwise
    \end{cases}
\end{alignedat}
\end{equation}

\noindent \textbf{Update ${\bf{F}}$:} When optimizing for ${\bf{F}}$, Eq. \eqref{eq:ALM} becomes
\begin{equation}
\label{eq:s12}
\begin{alignedat}{2}
\argmin_{{\bf{F}}} \frac{\mu}{2}||{\bf{X}}-{\bf{FG}}^T -{\bf{E}}+ \frac{1}{\mu}{\bf{\Lambda}}_1||_F ^2
\end{alignedat}
\end{equation}
Hence we obtain the solution  $${\bf{F}}=({\bf{X}}-{\bf{E}}+\frac{1}{\mu}{\bf{\Lambda}}_1){\bf{G}}({\bf G}^T{\bf G})^{-1}$$.

\noindent \textbf{Update ${\bf{H}}$:} When optimizing for ${\bf{H}}$, Eq. \eqref{eq:ALM} becomes
\begin{equation}
\label{eq:s16}
\begin{alignedat}{2}
\argmin_{{\bf{H}}\geqslant 0} \frac{\mu}{2}||{\bf{G}}-{\bf{H}} + \frac{1}{\mu}{\bf{\Lambda}}_2||_F ^2 +  \lambda \upt \upr {\bf{G}}^T {\bf{L_Z H}}
\end{alignedat}
\end{equation}
which can be re-arranged to the following form:

\begin{equation}
\label{eq:s17}
\begin{alignedat}{2}
\argmin_{{\bf{H}}\geqslant 0} ||{\bf{H}}-{\bf{J}}||_F ^2
\end{alignedat}
\end{equation}
where ${\bf{J}}= ({\bf{G}}+\frac{1}{\mu}{\bf{\Lambda}}_2 -\frac{\lambda}{\mu}{\bf{L_Z}}{\bf G})$. 
The solution for ${\bf{H}}$ can be obtained as
\begin{equation}
\label{eq:s19}
\begin{alignedat}{2}
{\bf{H}}_{ij}= max({\bf{J}}_{ij},0), i=1,2,\dots,n, ~j=1,2,\dots,c.
\end{alignedat}
\end{equation}

\noindent \textbf{Update ${\bf{G}}$:} When optimizing with respect to ${\bf{G}}$, Eq. \eqref{eq:ALM} becomes
\begin{equation}
\label{eq:s6}
\begin{alignedat}{2}
\argmin_{{\bf{G}}^T{\bf{G}}={\bm I}} \frac{\mu}{2}||{\bf{X}}-{\bf{FG}}^T-{\bf{E}}+\frac{1}{\mu}{\bf{\Lambda}}_1||_F ^2 \\+\frac{\mu}{2}||{\bf{G}}-{\bf{H}}+\frac{1}{\mu}{\bf{\Lambda}}_2||_F ^2 + \lambda \upt \upr {\bf{G}}^T {\bf{L_Z H}}\\
\end{alignedat}
\end{equation}
This can be further simplified to
\begin{equation}
\label{eq:s17}
\begin{alignedat}{2}
\argmin_{{\bf{G}}^T{\bf{G}}={\bm I}} ||{\bf{G}}-{\bf{N}}||_F ^2
\end{alignedat}
\end{equation}
\begin{equation}
\label{eq:s8}
\begin{alignedat}{2}
{\bf{N}}= {\bf{H}} - \frac{1}{\mu}{\bf{\Lambda}}_2 - \frac{\lambda}{\mu}{\bf{L_Z H}} + ({\bf{X}}-{\bf{E}} + \frac{1}{\mu}{\bf{\Lambda}}_1)^T {\bf{F}}
\end{alignedat}
\end{equation}
The solution to \eqref{eq:s8} is:
\begin{equation}
\label{eq:s11}
\begin{alignedat}{2}
{\bf{G}}={\bf{UV}}^T
\end{alignedat}
\end{equation}
where ${\bf{U}}$ and ${\bf{V}}$ are left and right singular values of the singular value decomposition of ${\bf{N}}$.\\

\section{ Algorithms.}
Algorithm \ref{alg:A2DM2} describes the pseudo code to apply Accelerated ADMM (A2DM2) on DS3 to initialize variable Z. This algorithm solves equation (4.15) in the paper. Algorithm \ref{alg:FAST} describes the pseudo code of a convex optimization to solve for variable ${\bf Z}$ in f-SMRMF. This algorithm solves equation (5.18) in the paper.
\begin{algorithm}[t]  
    \caption{Accelerated ADMM (A2DM2) to solve (4.15) in the paper}
    \label{alg:A2DM2}
    \begin{algorithmic}[1]
   \STATE Initialize: $\Hat{{\bf{C}}}^0,\Hat{{\bf{\Lambda}}}^0 = {\bf{0}}, a_0=1, \mu=10^{-1}, \rho=1.05 $ \\
    \REPEAT
    \STATE ${\bf{Z}}^{t} =\argmin_{\bf{Z}}  L({\bf{Z}},\Hat{{\bf{C}}^t},\Hat{{\bf{\Lambda}}}^t)$\\
   \STATE  ${\bf{C}}^{t} =\argmin_{\bf{C}}  L({\bf{Z}}^t,{\bf{C}},\Hat{{\bf{\Lambda}}}^t)$\\
   \STATE  ${\bf{\Lambda}}^t = \Hat{{\bf{\Lambda}}}^t + \mu({\bf{Z}}^{t} - {\bf{C}}^{t}) $\\
   \STATE  $a_{t+1} = \frac{1+\sqrt{1+4a_t^2}}{2}$\\
   \STATE  $\Hat{{\bf{\Lambda}}}^{t+1} = {\bf{\Lambda}}^t + \frac{a_t - 1}{a_{t+1}} ({\bf{\Lambda}}^t - {\bf{\Lambda}}^{t-1})$\\
    \STATE $\Hat{{\bf{C}}}^{t+1} = \argmin_{{\bf{C}}} tr({\bf{D}}^T {\bf{C}}) + \frac{\delta}{2} || {\bf{C}} ||_F^{2} +\langle {\Hat{\bf{\Lambda}}^{t+1}}, {\bf{Z}} - {\bf{C}} \rangle ~\ups.\upt. ~{\bf{1}}^T {\bf{C}} = {\bf{1}}^T; {\bf{C}} \geq 0$
    \STATE $\mu = \rho\mu$
     \UNTIL{convergence}
   \end{algorithmic}
   \end{algorithm}
\begin{algorithm}[t]  
   \caption{Algorithm to solve (5.18) in the paper}
    \label{alg:FAST}
    \begin{algorithmic}[1]
   \STATE Initialize: $z^0 = 0,$ $s^0=1$, $m=1+\max_{1\le i\le n}d_{ij},$ $d =d_j,$ $l_0=\min_{1\le i\le n}d_{i},$ $i_0 = \argmin_{1\le i\le n}d_{i}, $
    $d_{i_0} = m,$ $I_a^0 =\{i_0\}. $\\
    \STATE $r = l_0+\delta.$ 
    \WHILE{$r>l_0$ and $t<n-1$} 
    \STATE $l_1=\min_{1\le i\le n}d_{i},$ $i_1 = \argmin_{1\le i\le n}d_{i}, $
    $d_{i_1} = m.$\\
    \STATE $w = (l_1-l_0)/\delta.$
    \FOR{$i\in I_a^{t-1}$} \STATE $z_i^{t} = z_i^{t-1}+w,$ $s^{t} =s^{t-1}-w $ \ENDFOR
   \STATE  $l_0 = l_1,$ $I_a^{t} =I_a^{t-1}\cup i_1$ \\
   \STATE  $r=l_0+\delta s^t/|I_a^t|$\\
     \ENDWHILE
    \STATE  $z = z^t+s^t/|I_a^t|$\\

   \end{algorithmic}
   \end{algorithm}
   
\section{Empirical Evaluation.}
\subsection{Simulated Data.}
\begin{figure*}[t]
 \centering
      \subfigure[2DMoons (RMNMF)]{
        \centering
        \includegraphics[width=40mm,height=30mm]{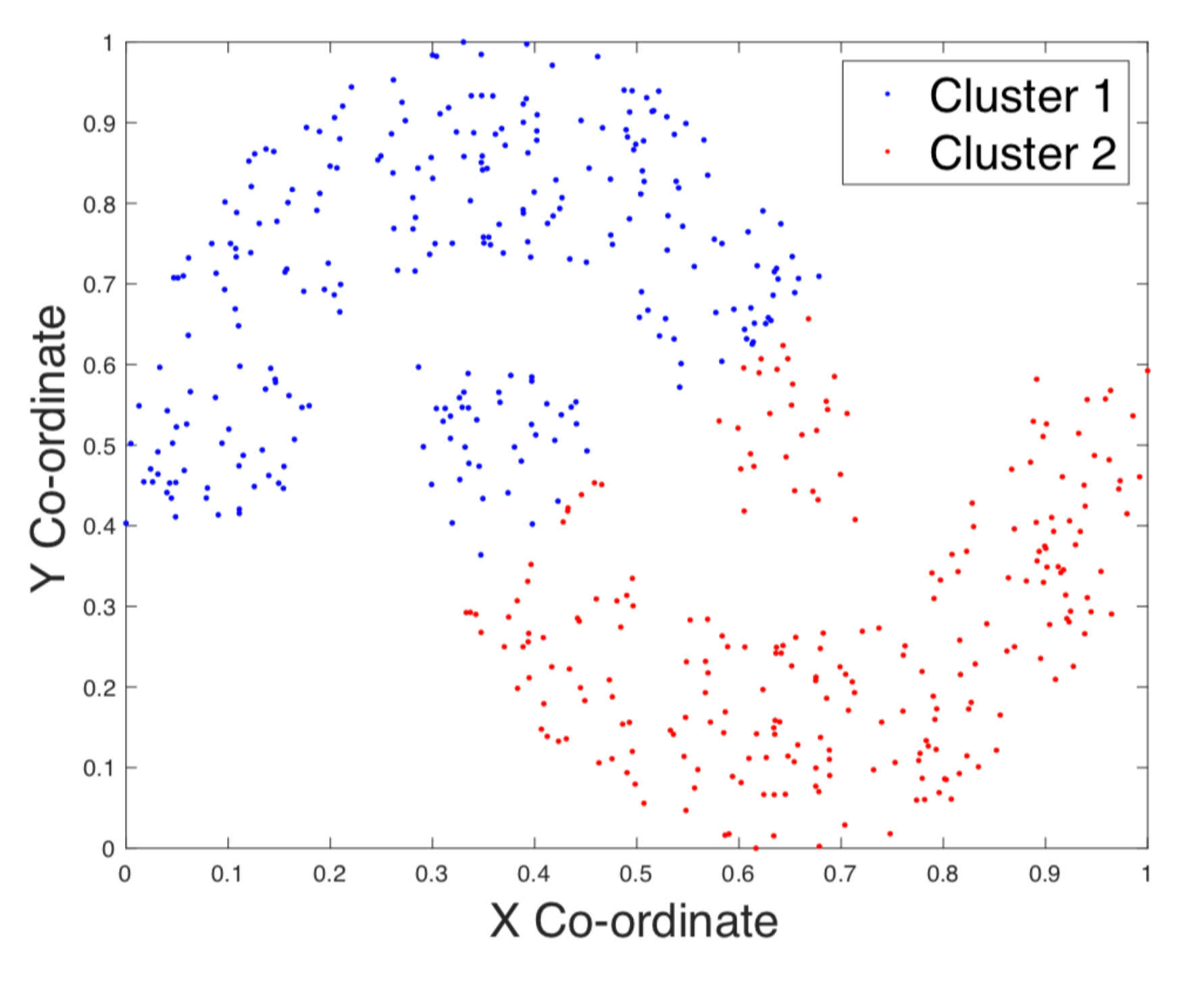}
        }
      \subfigure[2DMoons (SMRMF)]{
        \centering
        \includegraphics[width=40mm,height=30mm]{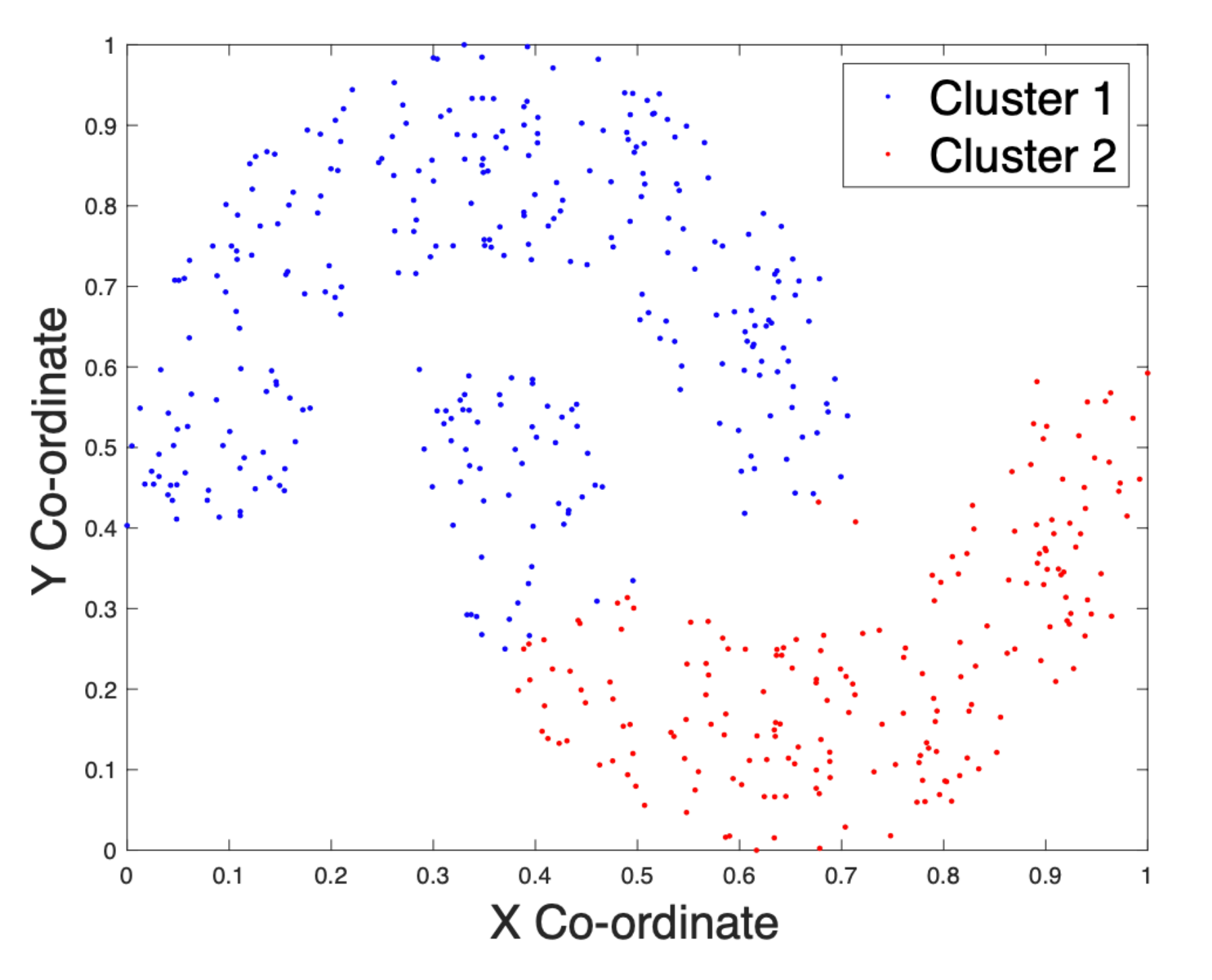}
        }
        \subfigure[10DMoons (RMNMF)]{
        \centering
        \includegraphics[width=40mm,height=30mm]{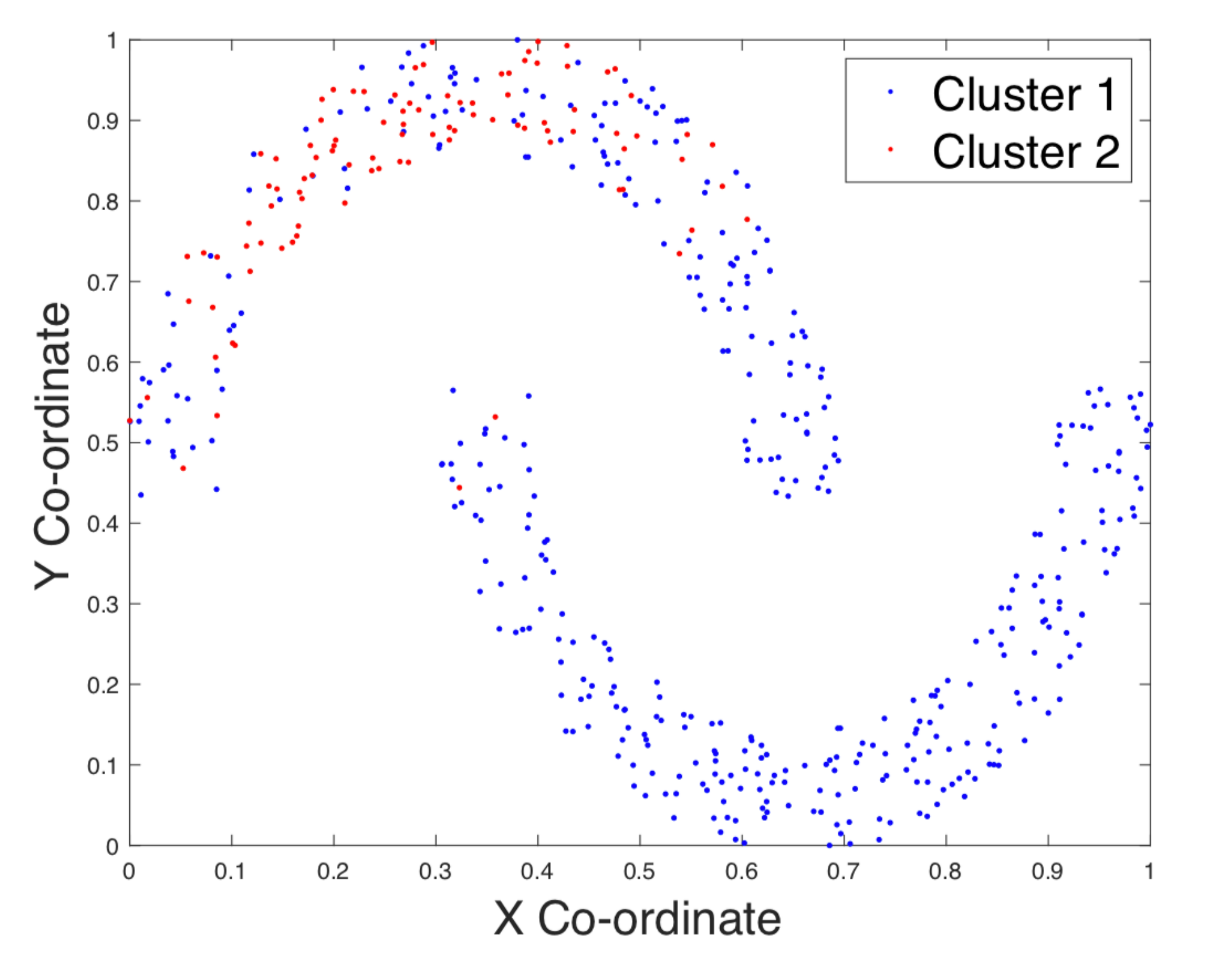}
        }
         \subfigure[10DMoons (SMRMF)]{
        \centering
        \includegraphics[width=40mm,height=30mm]{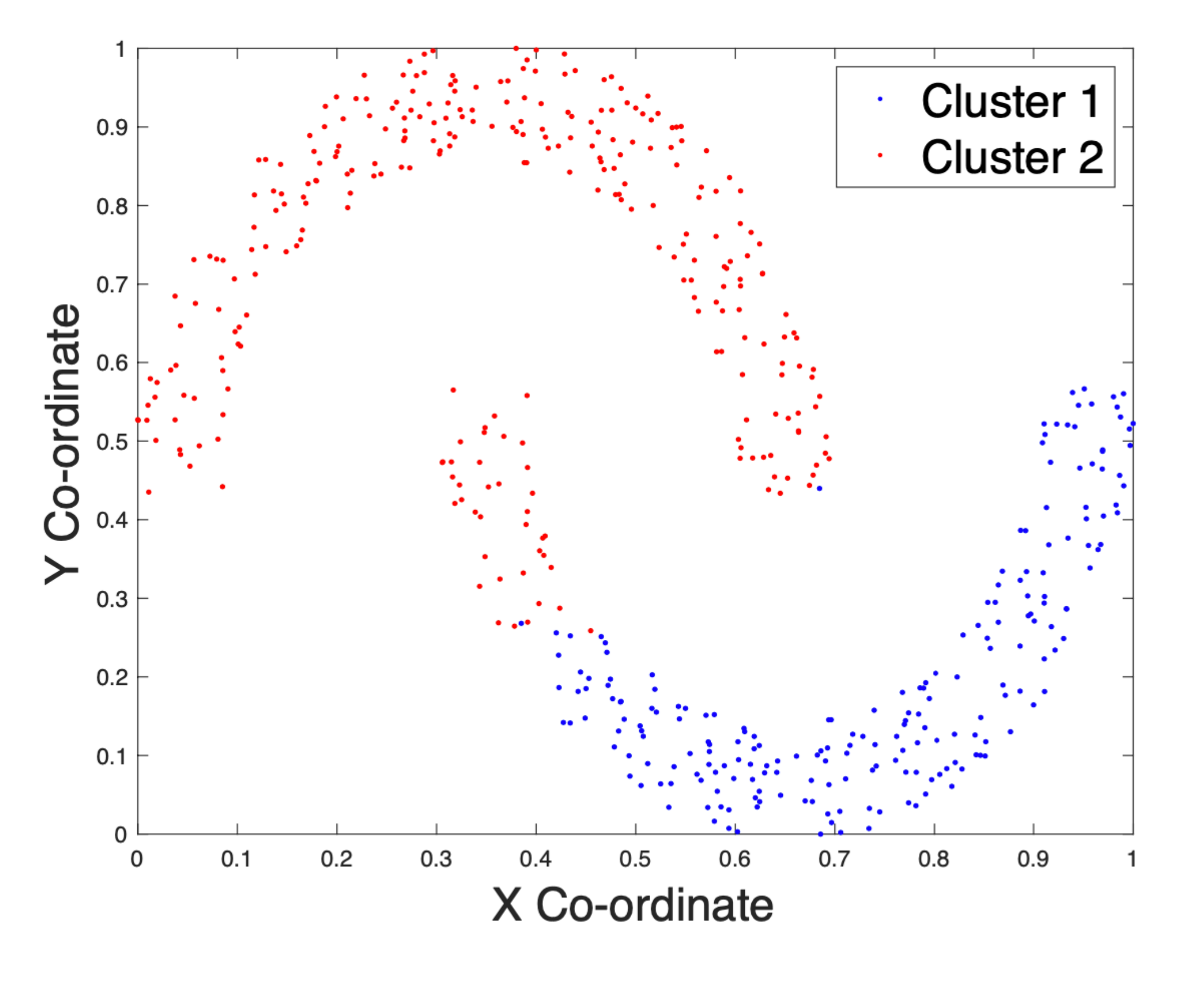}
        }
        \caption{Clustering of moons data.} 
        \vspace{-1.0em}
    \label{fig:manifold}
\end{figure*}

The data clustering of moons data by RMNMF and SMRMF are given in Fig \ref{fig:manifold}. The data is projected on the first two dimensions to plot 10DMoons.

\subsection{Data Sets.}
\label{sec:EMPEVALDATA}
This section describes the datasets used in Table 3 in the paper. We chose real data sets with sufficient variations in density, class distribution, and dimensionality. Except for ORL\footnote{www.cad.zju.edu.cn/home/dengcai/Data/FaceData.html}, OVA\textunderscore Colon\footnote{www.openml.org/d/1161}, and 20-news\footnote{www.cad.zju.edu.cn/home/dengcai/Data/TextData.html}, the remaining data sets are obtained from the UCI Machine Learning Repository \cite{UCI}. COIL originally consists of 100 classes, out of which the first 5 classes are selected, with 72 images in each class. For ISOLET, the training data (isolet1+2+3+4 in the UCI repository) of five classes (A,B,C,D,E) are selected. 
The first 200 samples are selected from each class in Wave. The attributes `musk name' and `conformation name' are omitted for Musk1.  An equal number of instances from each class is chosen for 20-newsgroups, and min and max document frequency thresholding is applied to remove rare and very frequent words. The data is then represented using TF-IDF values. The features of each data set are scaled to the range [0, 1] using min-max scaling. Instances are normalized to unit $L_2$ norm for each data set. Duplicate instances and invariant features are removed from each data set.

\subsection{Parameter Setting for Comparison Experiments.}
We tune the regularization parameters of the algorithms 
in the range $\left\{10^{-3}, 10^{-2}, \dots, 1\right\}$ and report the best results. For APMF, the regularization parameters for the higher-dimensional datasets (ISOLET, COIL-5, 20-newsgroups, ORL, and OVA\textunderscore Colon) are set to 0.01, as suggested by the authors (APMF on these data sets did not run in a reasonable amount of time to enable fine-tuning). The parameter $\alpha$ for AMRMF is set according to the derived result in \cite{CAN}. The parameter $\tau$ for SMRMF is set as $0.1 \times n$, where $n$ is the number of points, so that 10\% of the data is selected as exemplars. The parameter $\delta$ in A2DM2 initialization is set to 0.01, while $\delta$ in SMRMF(fast) is tuned in the range $\left\{10^{-2}, 10^{-1}, \dots, 10\right\}$. The ALM parameters for all experiments are initialized as follows: ${\bf \Lambda}_1 = {\bf \Lambda}_2 = {\bf \Lambda}_3 = \mathbf{0}$, $\rho=1.05$, and the maximum iterations is set to 1000. The value of $k$ for $k$-NN graph construction is set to $\sqrt{n}$ for all algorithms. The number of clusters, which is equal to the rank of the factor matrices, is set equal to the number of classes in the data, as done in literature (RMNMF, AMRMF, APMF, ESC-FFS).  
For DS3 based clustering, the number of exemplars is set equal to the number of classes. The exemplar size for DPP and IPM are also set to 10\%. A dual DPP formulation is used in order to scale to larger datasets. The Gram matrix for DPP is computed using pairwise cosine similarities. The reduced data dimensionality for APMF is selected from three parameter settings: $\left\{0.2m, 0.5m, 0.7m \right\}$, where $m$ is the full dimensionality of the data. The parameter $\lambda$ for ESC-FFS is set to 100 as suggested by the authors. The exemplar size for ESC-FFS is varied in the set $\left\{ 1\%, 2\%, 3\%, 4\%, 5\%, 10\%, 15\%\right\}$, and the best average result is reported. The neural network for DSMF has two hidden layers, as suggested by the authors. We set their sizes as $layer_1 = {m}/{2}$ and $layer_2=max(\frac{layer_1}{10},2 \times c)$, where $m$ is the data dimensionality and $c$ is the number of clusters. 
Except for DSMF and APMF, all the matrix factorization algorithms are run until the following convergence criterion is met: ${|J^{t+1}-J^{t}|}/{J^t} < 10^{-4}$, where $J^{t}$ is the objective function value for an algorithm at time $t$. For DSMF and APMF, we follow the convergence criterion in the available code. GNMF, RMNMF, APMF, and SMRMF construct a $k$-NN graph from the affinity matrix, while AMRMF and SMRMF-NS use a fully connected graph for regularization. The pairwise affinities of the $k$-NN graph of RMNMF are computed using an adaptive heat kernel:
 $A({\bf{x}}_i, {\bf{x}}_j) = e^{-{||{\bf{x}}_i - {\bf{x}}_j||^2}/{2\sigma_i^2}}$;
 where $\sigma_i$ is the distance of ${\bf{x}}_i$ to the $k^{th}$ nearest neighbor. The pairwise affinities for AMRMF, APMF, SMRMF, and its variants are learned. For SMRMF and its variants on real data, the adaptive bandwidths ($\sigma_i$ and $\gamma_i$ (in equation 4.10 of the paper) are set equal to the squared distance to the $k^{th}$ neighbor of a data point. f-SMRMF further optimizes run-time by setting adaptive bandwidths $\gamma_i$ as the mean of the adaptive bandwidths $\sigma_i$, in each iteration. In order to reduce the running time of SMRMF and variants for larger data sets ($n \geq 2000)$, we allow an early termination of the update of the variables $\bf{Z}$ and $\bf{C}$; they are updated until they satisfy
 $||{\bf Z}^{t+1} - {\bf Z}^{t}||_{\infty} \leq 1e^{-5}$ and
 $||{\bf C}^{t+1} - {\bf C}^{t}||_{\infty} \leq 1e^{-5}$, respectively.

\subsection{Running Time.}

Table~\ref{tab:RUNTIME} provides running time in seconds, for each data set on all the compared algorithms.


\begin{table*}[t]
\caption{Running times (in seconds) of all methods (standard deviations are given in parentheses).}
\label{tab:RUNTIME}
\begin{center}
\hspace{-35cm}
\vskip -0.15in
\begin{tiny}
\begin{tabular}{p{1.1cm}p{1.3cm}p{0.5cm}p{1.3cm}p{0.7cm}p{0.7cm}p{0.5cm}p{1.2cm}p{0.5cm}p{1.3cm}p{0.5cm}p{0.5cm}p{0.7cm}p{0.5cm}p{1.1cm}}
\toprule
 Data & NMF & DS3 & GNMF &  RMNMF & AMRMF & APMF &  DPP & IPM  & ESC-FFS & DSMF & SMRMF & SMRMF  & SMRMF & f-SMRMF \\
& & & & & & &  & & & &-Euc & -NS  & &  \\
 \midrule
 Wave & {\bf 0.09} (0.01) & 4.03 & 0.20 (0.01) & 0.88 & 5.34 & 6.36 & 0.76 (0.03) & 1.17 & 0.09 (0.004) & 1.37 & 1.11 & 61.55 & 100.80  & 0.97\\
 Ionosphere & {\bf 0.03} (0.001) & 0.32 & 0.06 (0.002) &  0.59 & 2.00 & 1.23 & 0.12 (0.01) & 0.61 & 0.04 (0.003) & 0.72 & 0.49 &  3.83 &  4.07 & 0.54\\
 Sonar & 0.37 (0.002) & 0.14 & 0.05 (0.002) & 0.45 & 0.89 & 1.69 & 0.47 (0.01) & 0.32 & {\bf 0.03} (0.001) & 0.81 & 0.20 & 0.69 & 0.88 & 1.01 \\
 Movement & 0.12 (0.02) & 0.28 & 0.23 (0.02) & 1.38 & 1.60 & 2.51 & 1.22 (0.06) & 1.76 & {\bf 0.05} (0.002) & 1.63 & 1.93 & 2.96 & 3.13 & 1.65\\
 Musk1 & {\bf 0.07} (0.003) & 0.44 & 0.10 (0.003) & 1.35 & 5.85 & 5.32 & 1.32 (0.05) & 1.70 & 0.08 (0.004) & 2.10 & 1.23 & 3.38 &  3.79 & 2.36\\
 mfeat-fac & {\bf 0.55} (0.02) & 9.84 & 1.52 (0.02) & 10.78 & 102.50 & 113.38 & 8.60 (0.2) & 34.90 & 0.77 (0.02) & 8.96 & 91.73 & 154.35 & 919.86 & 110.79\\
 mfeat-pix &  {\bf 0.62} (0.01) & 26.86 & 1.57 (0.03) & 11.02 &  151.81 & 288.49 & 9.68 (0.1) & 35.45 & 1.22 (0.03) & 8.63 & 108.66 & 4.63 & 569.56 & 72.51\\
Semeion & {\bf 0.58} (0.01) & 30.84 & 1.27 (0.02) & 8.55 & 44.57 & 68.14 & 10.82 (0.6) & 24.41 & 1.10 (0.03) & 8.51  & 25.71 & 777.90 & 933.61 & 195.71 \\
 ISOLET & 0.73 (0.01) & 2.70 &  1.09 (0.01) & 8.99 & 15.42 & 224.42 & 9.58 (0.2) & 16.18 & {\bf 0.40} (0.02) & 31.28 & 17.20 & 39.78 & 251.36 & 12.81 \\
 COIL-5 & 0.36 (0.01) & 0.66 & 0.39 (0.01) & 4.06 & 1.39 & 365.88 & 4.49 (0.1) & 5.74 & {\bf 0.10} (0.007) & 28.65 & 2.15 & 2.39 & 2.71 & 1.57\\
 20-news & {\bf 13.95} (0.2) & 182.22 & 21.12 (0.10) & 18.59 & 379.75 & 6447.00 & - & 1469.86 &  38.17 (0.68) & 651.76 & 888.42 & 621.95 & 642.90 & 697.20\\
 ORL & 6.73 (0.08) & 1.32 & 6.90 (0.14)  & 26.09 & 7.81 & 2947.20 & 48.05 (2.5) & 57.80 & {\bf 0.23} (0.02) & 92.50   & 7.16 & 47.60 & 47.86 & 8.27 \\
 OVA\textunderscore Colon & 15.01 (0.1) & 39.24 & 16.97 (0.04) & 34.46 & 155.07 & - & - & 1684.63 & {\bf 5.96} (0.4) & 2793.00 & 48.82 &  166.89 & 321.61 & 139.72\\
 \hline
 AVG & {\bf 3.02} & 22.99 & 3.96 & 9.78 & 67.23 & 872.64 & 8.64 & 256.50 & 3.71 & 279.23 & 91.91 &  148.45 & 292.47 & 95.77\\
\bottomrule
\end{tabular}
\end{tiny}
\end{center}
\vspace{-1em}
\end{table*}
\subsection{Analysis.}
\begin{figure}[th]
\begin{center}
\centerline{\includegraphics[width=\columnwidth]{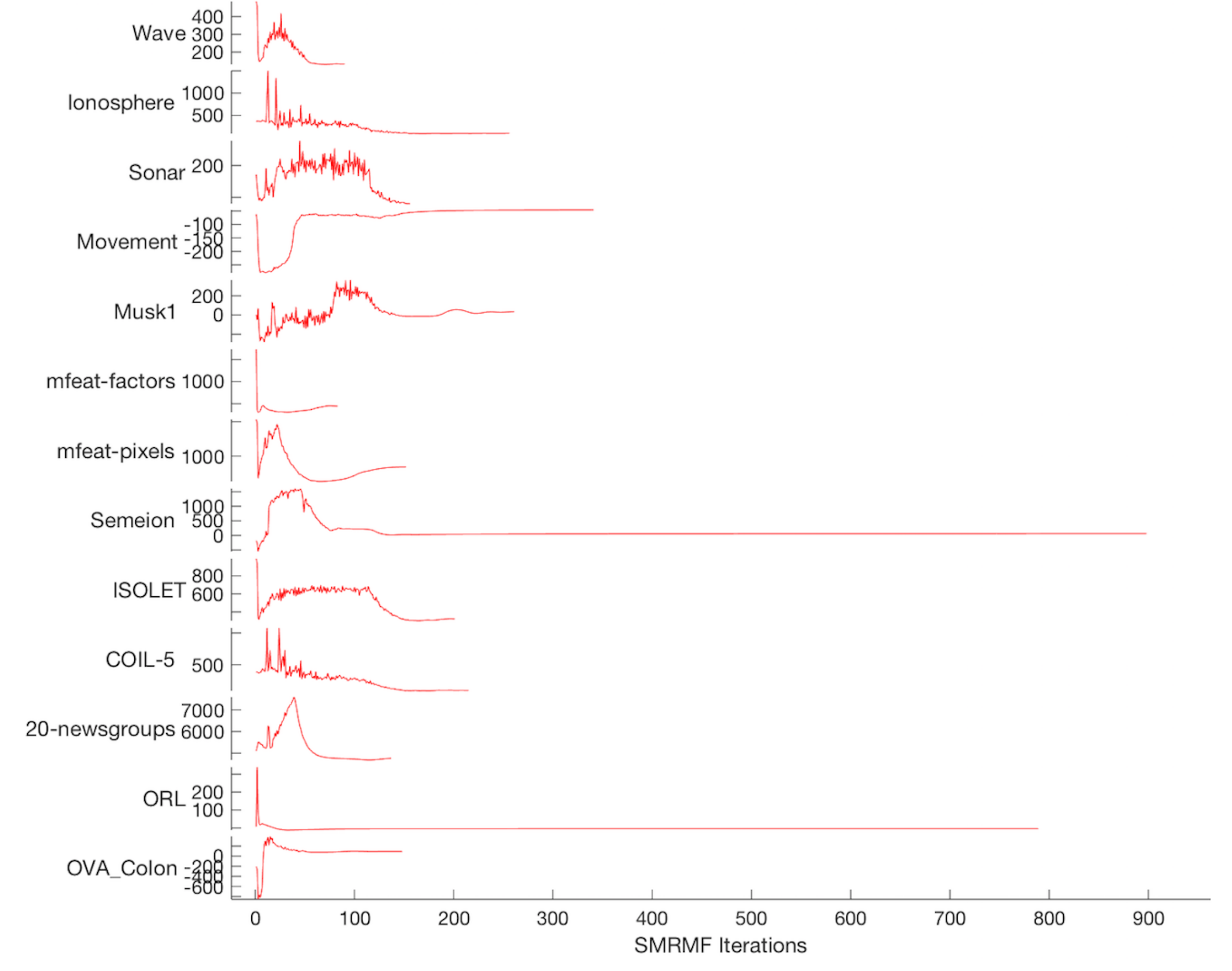}}
\end{center}
\vspace{-1em}
\caption{Convergence of the objective function.}
\vspace{-1em}
\label{fig:convergence}
\end{figure}

Here we analyze the convergence of our  approach, and its sensitivity to the number of nearest neighbors $k$ and to the exemplar set size.  Fig.~\ref{fig:convergence} shows the convergence of the unified objective function of SMRMF. The objective function converges in less than 1000 iterations for all data sets, with the majority of the data sets requiring fewer then 400 iterations.

\begin{figure}[th]
\begin{center}
\begin{small}
\centerline{\includegraphics[width=3in]{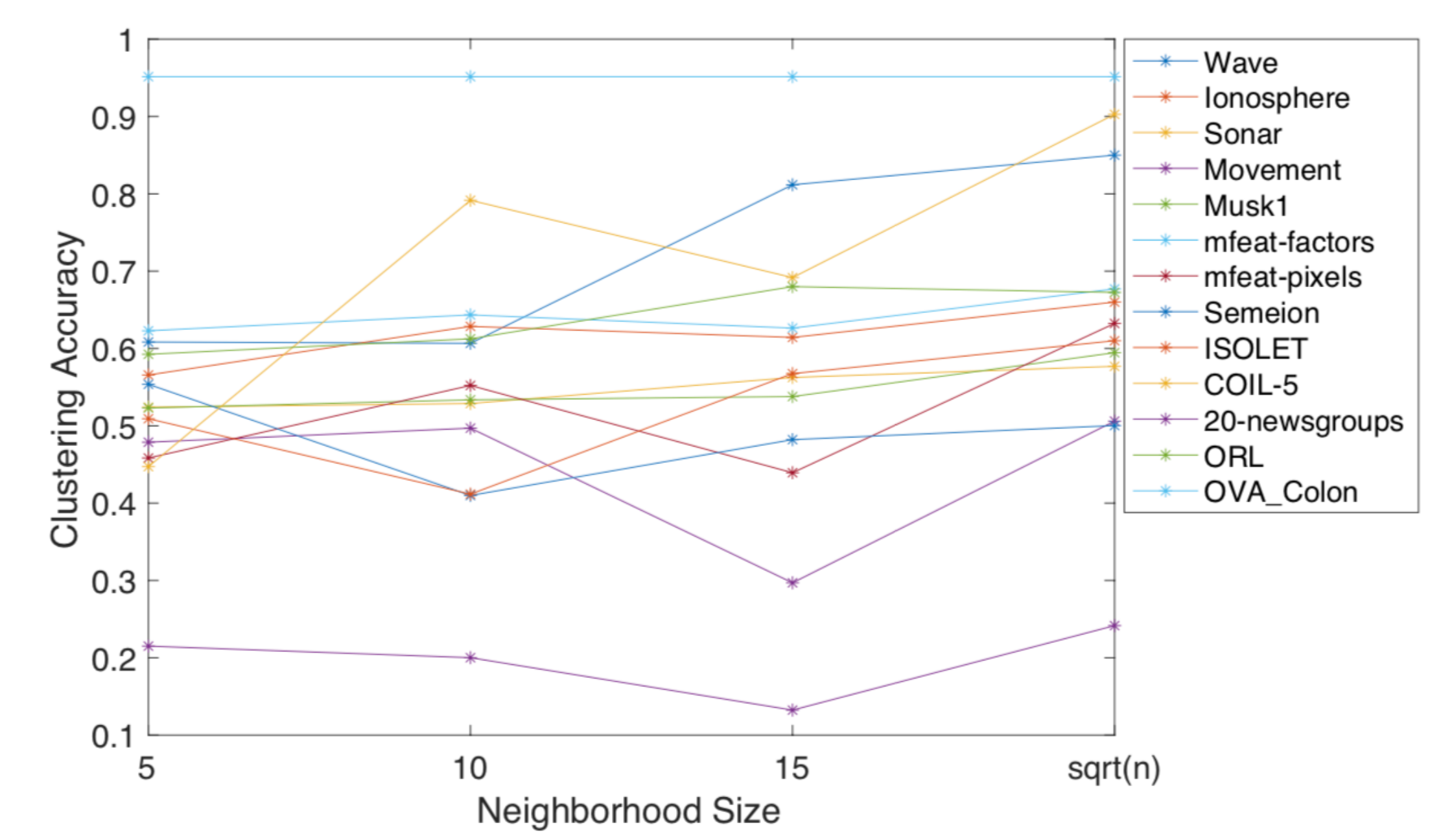}}
\end{small}
\end{center}
\vspace{-2em}
\caption{Sensitivity on $k$.}
\label{fig:kPlot}
\end{figure}

\begin{figure}[t!]
\begin{center}
\centerline{\includegraphics[width=3in]{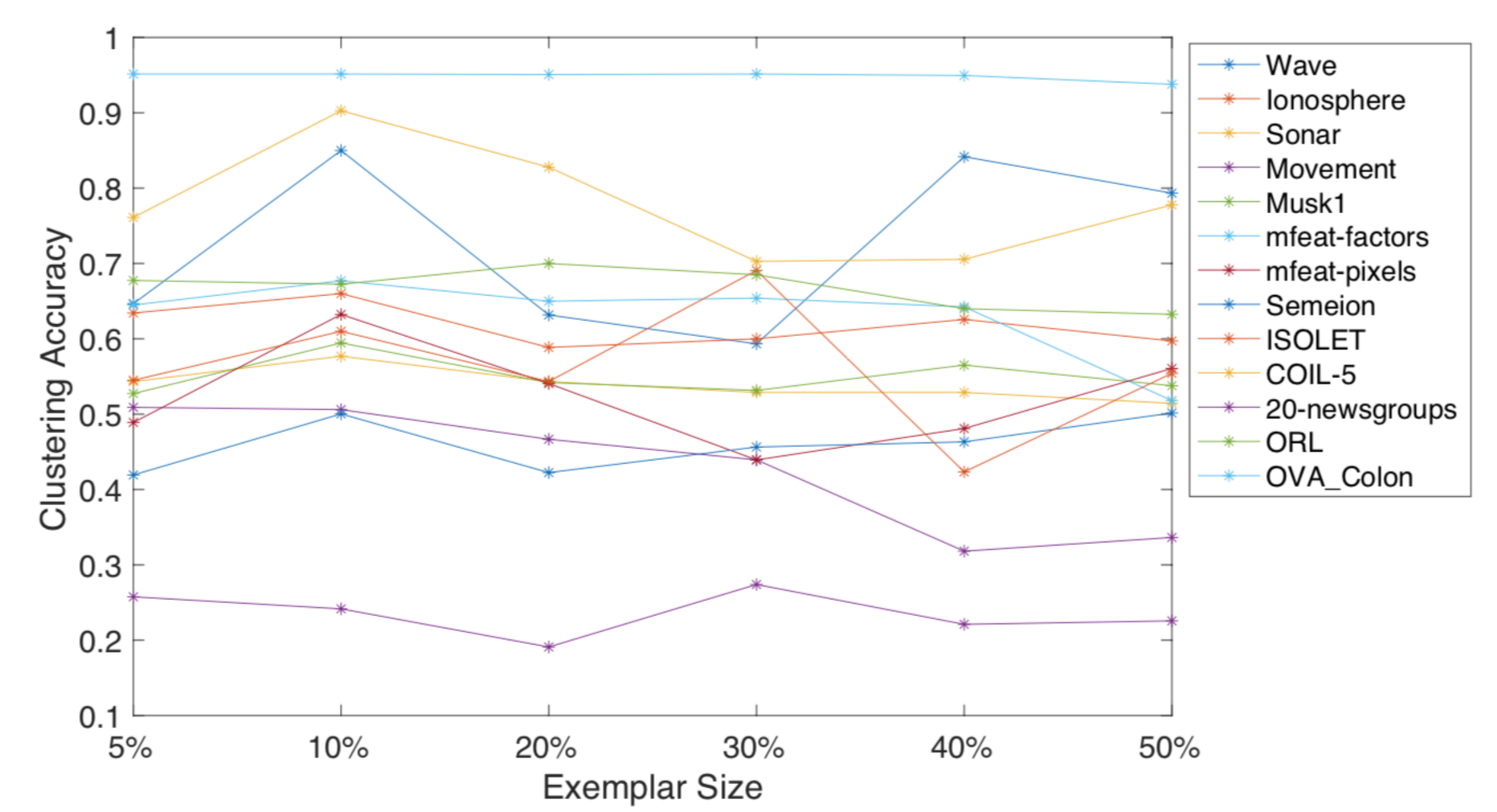}}
\end{center}
\vspace{-2em}
\caption{Sensitivity on exemplar set size.}
\vspace{-1em}
\label{fig:exemplarSize}
\end{figure}

Fig.~\ref{fig:kPlot} analyzes the sensitivity of accuracy on the neighborhood size $k$ for SMRMF. $k$ varies in the set $\left\{5, 10, 15, \sqrt{n}\right\}$, where $n$ is the number of points. We observe that $k=\sqrt{n}$ gives the best  accuracy for the majority of the data sets. Note that for Sonar, $\sqrt{n} < 15$.
Fig.~\ref{fig:exemplarSize} shows the effect of the exemplar set size on  accuracy. The exemplar size (\%) varies in the set $\left\{0.1, 0.2, \dots, 0.5\right\}$. Most data sets reach the highest  accuracy at exemplar set size $0.1$, and consequently manifest an overall slight decreasing trend. 

